\documentclass[10pt,technote,onecolumn]{IEEEtran}
\usepackage{cite}

\ifCLASSINFOpdf
   \usepackage[pdftex]{graphicx}
   \DeclareGraphicsExtensions{.pdf,.jpeg,.png,.eps}
\else
\fi
\usepackage{amsmath,amsthm,amssymb,amsfonts,bbm,mathrsfs}
\newtheorem{lemma}{Lemma}
\newtheorem{definition}{Definition}
\newtheorem{corollary}{Corollary}
\newtheorem{assumption}{Assumption}

\newtheorem*{proof*}{Proof}
\newtheorem*{remark*}{Remark}

\usepackage{algorithmic,algorithm}

\usepackage{array}

\usepackage{multirow,multicol}
\usepackage{color,xcolor}
\usepackage{setspace}
\usepackage{caption,subcaption}
\usepackage{booktabs}
\usepackage{diagbox}

\usepackage{url}

\begin{document}
%
\title{Robust Low-Rank Tensor Ring Completion}
%
%
%

\author{Huyan~Huang,
Yipeng~Liu,~\IEEEmembership{Senior Member,~IEEE},
Ce~Zhu,~\IEEEmembership{Fellow,~IEEE}
\thanks{This research is supported by National Natural Science Foundation of China (NSFC, No. 61602091, No. 61571102) and Sichuan Science and Technology Program (No. 2019YFH0008, No.2018JY0035).}        
\thanks{All the authors are with School of Information and Communication Engineering, University of Electronic Science and Technology of China (UESTC), Chengdu, 611731, China. (email: huyanhuang@gmail.com, yipengliu@uestc.edu.cn, zhenlong@std.uestc.edu.cn. eczhu@uestc.edu.cn).}
}

%
%

\markboth{Journal of \LaTeX\ Class Files,~Vol.~XX, No.~X, Month~Year}
{Shell \MakeLowercase{\textit{et al.}}: Bare Demo of IEEEtran.cls for IEEE Journals}
%



\maketitle

\begin{abstract}

Low-rank tensor completion recovers missing entries based on different tensor decompositions. Due to its outstanding performance in exploiting some higher-order data structure, low rank tensor ring has been applied in tensor completion. To further deal with its sensitivity to sparse component as it does in tensor principle component analysis, we propose robust tensor ring completion (RTRC), which separates latent low-rank tensor component from  sparse component with limited number of measurements. The low rank tensor component is constrained by the weighted sum of nuclear norms of its balanced unfoldings, while the sparse component is regularized by its $\ell_1$ norm. We analyze the RTRC model and gives the exact recovery guarantee. The alternating direction method of multipliers is used to divide the problem into several sub-problems with fast solutions. In numerical experiments, we verify the recovery condition of the proposed method on synthetic data, and show the proposed method outperforms the state-of-the-art ones in terms of both accuracy and computational complexity in a number of real-world data based tasks, i.e., light-field image recovery, shadow removal in face images, and background extraction in color video.
\end{abstract}


%
\IEEEpeerreviewmaketitle

\section{Introduction}
%
%
%
%

\IEEEPARstart{I}{n} practice, acquired data can be incomplete and grossly corrupted, such as computed tomography \cite{xia2019spectral}, magnetic resonance imaging \cite{poddar2019manifold, lin2018efficient}, visual camera \cite{moore2019panoramic}, recommendation system \cite{udell2016generalized}, etc.

Robust matrix completion (RMC) simultaneously recovers the incomplete entries and extracts the latent low-rank component in observed data \cite{candes2011robust, ji2010robust, chen2011robust, klopp2017robust, cherapanamjeri2017nearly, nie2012robust, cambier2016robust, mansour2014video, jiang2017robust}. Assuming that we have observed a part of the data $\mathbf{T}$, and the observation can be denoted as $\mathbf{T}_\mathbb{O}$, where the set $ \mathbb{O} $ indicates which part of the data are observed. RMC can provide the low rank component $ \mathbf{L} $ and sparse component $ \mathbf{S} $ from the observation $\mathbf{T}_\mathbb{O}$, which can be formulated as the following optimization model:

\begin{align}
\label{eq: RMC}
\min_{\mathbf{L},~\mathbf{S}}\; \lVert \mathbf{L} \rVert_\text{mn}+\lambda\lVert \mathbf{S} \rVert_1,~\text{s. t.}~~\mathbf{L}_{\mathbb{O}}+\mathbf{S}_{\mathbb{O}}=\mathbf{T}_{\mathbb{O}}
\end{align}
where $ \lVert \mathbf{L} \rVert_\text{mn} $ is the matrix nuclear norm which sums all the singular value of the matrix $ \mathbf{L} $, and $ \lVert \mathbf{S} \rVert_1 $ is the $ \ell_1 $ norm of the matrix $ \mathbf{S} $ which sums all the absolute values of the entries, and $ \lambda $ is the parameter for making balance for these two terms.

RMC can only handle the data in matrix form, and higher-order signals are needed to be unfolded into matrices for processing. The subspace structure would be partly lost in such matricization, which leads to unsatisfactory performance for higher-order signals. 

Tensor is a natural representation for higher-order data, and robust tensor completion (RTC) is proposed to avoid such performance degeneration \cite{kolda2009tensor, cichocki2015tensor, cichocki2016tensor}. The optimization model of RTC can be formulated as follows:

\begin{align}
\label{eq: RTC}
 \min_{\mathcal{L},~\mathcal{S}} \lVert \mathcal{L} \rVert_*+\lambda \lVert \mathcal{S} \rVert_1,~\text{s. t.}~~\mathcal{L}_{\mathbb{O}}+\mathcal{S}_{\mathbb{O}}=\mathcal{T}_{\mathbb{O}}.
\end{align}
where $ \mathcal{T} $ is the higer-order data,  $ \mathcal{L} $ and $ \mathcal{S} $ are the low rank and sparse components, and $ \lVert \mathcal{L} \rVert_* $ denotes the tensor nuclear norm which has different definitions with different tensor decompositions, and $ \lVert \mathcal{S} \rVert_1 $ is the $\ell_1$ norm which sums all the absolute values of the entries in $ \mathcal{S} $.  

RTC can be regarded as a generalized form of tensor completion and robust tensor principal component analysis. Tensor completion recovers data from its limited samples based on the pre-defined low-rank tensor assumption, which can be formulated as follows \cite{liu2019low, long2018low, liu2019image}:
\begin{align}
\label{eq: TC}
 \min_{\mathcal{L},~\mathcal{S}}\; \lVert \mathcal{L} \rVert_*,\;\text{s. t.}\;\mathcal{L}_{\mathbb{O}}=\mathcal{T}_{\mathbb{O}}.
\end{align}
It can suffer the performance deterioration from gross corruption. Robust tensor principal component analysis separates low-rank tensor component from sparse noise, and the corresponding optimization model is as follows \cite{liu2018improved}:
\begin{align}
\label{eq: RTPCA}
 \min_{\mathcal{L},~\mathcal{S}}~~ \lVert \mathcal{L} \rVert_*+\lambda\lVert \mathcal{S} \rVert_1,~~\text{s. t.}~~\mathcal{L}+\mathcal{S}=\mathcal{T}. 
\end{align}
As we can see in the optimization model, it requires the full observation. As a more general form, RTC can reduce the sample complexity of robust tensor principal component analysis, and enhance the robustness of tensor completion. 

Existing RTC methods are based on different tensor decompositions \cite{huang2014provable, wang2019robust, carozzi2019robust, jiang2018low}. the Tucker decomposition based RTC method is proposed with exact recovery guarantee in \cite{huang2014provable}. The sum of nuclear norms of factor matrices is used as the suboptimal convex surrogate of the Tucker-rank $\operatorname{rank}_{\text{Tucker}}\left(\mathcal{X}\right)=\left[\operatorname{rank}\left(\mathbf{X}_{\left(1\right)}\right),\cdots,\operatorname{rank}\left(\mathbf{X}_{\left(D\right)}\right)\right]$, where $\mathbf{X}_{\left(d\right)}$ is the tensor unfolding along its $d$-th dimension \cite{kolda2009tensor}. Tensor singular value decomposition (t-SVD) factorizes a $3$-way tensor into two orthogonal tensors and a f-diagonal tensor based on the tensor-tensor product \cite{ kilmer2013third}, and the corresponding tubal rank is defined as the number of non-vanishing tubes in the f-diagonal tensor \cite{lu2019tensor}. The t-SVD based RTC method is proposed in \cite{wang2019robust} with strong recovery guarantee of non-asymptotic upper bounds on the estimation error. In addition, some other RTC methods based on parallel matrix factorization and canonical polyadic (CP) decomposition are given in \cite{carozzi2019robust, jiang2018low}.  

The recently proposed tensor ring (TR) decomposition factorizes a high-order tensor into a sequence of cyclically contracted $3$-order tensors \cite{zhao2017learning, ye2018tensor}. The TR rank is consistently invariant under the cyclic permutation of the factors, which induces the circular TR unfolding scheme \cite{huang2020provable}, and TR is more powerful in low rank representation than some other decompositions in a number of applications \cite{bengua2017efficient, wang2017efficient}. 
Therefore, in this paper, we employ TR for RTC to enhance it performance in applications. 

Motivated by the recovery guarantee for tensor completion developed in \cite{huang2020provable}, in this paper we construct a novel RTC model via TR decomposition, extend the recovery guarantee to RTC, and apply it to some image reconstruction applications. Considering the modes of practical data often have different correlations and dimensional sizes, we employ multiple circular unfoldings and propose a weight sum of nuclear norm model which can automatically tune the weights according to the estimated sample complexity. The alternating direction method of multipliers (ADMM) divides the optimization model into several sub-problems that can be fast solved. Numerical experiments on synthetic data verify our theoretical analysis, and the experiments on real-world data such as light field images, YaleB face dataset and color videos illustrate the proposed method's superiority over the existing ones in terms of both computational complexity and recovery accuracy.

The rest parts of this paper is organized as follows. In section \ref{section-notation}, the basic notations of TR are introduced. In section \ref{section-method}, we present the robust tensor ring completion and analyze the exact recovery guarantee. Section \ref{section-experiment} displays the experimental results. Finally we conclude our work in section \ref{section-conclusion}.

\section{Notation and Preliminary}
\label{section-notation}

\subsection{Notation}

There are some basic notations of tensors. A scalar, a vector, a matrix and a tensor are denoted by normal letter, boldface lowercase letter, boldface uppercase letter and calligraphic letter, respectively. For instance, a $D$-order tensor is denoted as $\mathcal{X} \in \mathbb{R}^{N_{1}\times \dotsm  \times N_{D}}$, where $N_d$ is the size corresponding to dimension $ d,\; d \in \left\{1,\dotsc,D\right\}$. An entry of the tensor $ \mathcal{X} $ is denoted as $\mathcal{X}\left(n_1,\dotsc,n_D\right)$, where $n_d$ is the index with mode $d,\; 1\leq n_{d}\leq N_{d}$. A mode-$d$ fiber of $\mathcal{X}$ can be denoted as $\mathcal{X}\left(n_1,\dotsc,n_{d-1},:,n_{d+1},\dotsc,n_D\right)$, and $\mathcal{X}\left(\dotsc,n_d,\dotsc\right)$ represents the slice along mode $d$.

The Kronecker product and Hadamard product are denoted by $\otimes$, $\circledast$ respectively. $\mathcal{I}$ is a tensor whose elements are all $1$ and $\mathscr{E}: \mathcal{X} \mapsto \mathcal{X}$ is the identity operator. For instance, $\mathscr{E}\left(\mathcal{X}\right)=\mathcal{I}\circledast \mathcal{X}$. The inner product of $\mathcal{X}$ and $\mathcal{Y}$ is defined as $\langle \mathcal{X},\mathcal{Y} \rangle=\sum_{n_1=1}^{N_1}\dotsm \sum_{n_D=1}^{N_D}\mathcal{X}\left(n_1,\dotsc,n_D\right)\mathcal{Y}\left(n_1,\dotsc,n_D\right)$. The Frobenius norm of $\mathcal{X}$ is defined as $\lVert \mathcal{X} \rVert_\mathrm{F}=\sqrt{\langle \mathcal{X},\mathcal{X} \rangle}$. The $\ell_1$ norm of $\mathcal{X}$ is defined as $\langle \mathcal{I},|\mathcal{X}| \rangle$, where $|\cdot|$ means the absolute value.

\subsection{Preliminary on Tensor Ring Decomposition}

\begin{definition}[TR decomposition \cite{zhao2017learning, ye2018tensor}]
Letting $\mathcal{U}_1$, $\dotsc$, $\mathcal{U}_D$ denote the TR factors, where $\mathcal{U}_d \in \mathbb{R}^{R_d\times N_d \times R_{d+1}}$ with the TR rank $\left[R_1,\dotsc,R_D\right]$, the TR decomposition is defined as
\begin{equation}
\label{eq: TR decomposition}
\begin{split}
\mathcal{X}=& \sum_{r_1=1}^{R_1}\dotsm \sum_{r_D=1}^{R_D}\mathcal{U}_1\left(r_1,:,r_2\right)\circ \dotsm \circ \\
& \mathcal{U}_{D-1}\left(r_{D-1},:,r_D\right)\circ \mathcal{U}_D\left(r_D,:,r_1\right)
\end{split}
\end{equation}
where $\circ$ denotes the outer product.

\end{definition}
In addition, (\ref{eq: TR decomposition}) has two equivalent definitions in  scalar form as follows:
\begin{equation}
\begin{split}
\mathcal{X}\left(n_1,\dotsc,n_D\right)=& \sum_{r_1=1}^{R_1}\dotsm \sum_{r_D=1}^{R_D}\mathcal{U}_1\left(r_1,n_1,r_2\right) \dotsm \\
& \mathcal{U}_{D-1}\left(r_{D-1},n_{D-1},r_D\right) \mathcal{U}_D\left(r_D,n_D,r_1\right) \\
=& \operatorname{tr}\left( \mathcal{U}_1\left(:,n_1,:\right)\dotsm \mathcal{U}_D\left(:,n_D,:\right) \right),
\end{split}
\end{equation}
where $\operatorname{tr}\left(\cdot\right) $ is the trace function.

\begin{definition}[Tensor ring connection product \cite{wang2017efficient}]
Let $\overline{\otimes}$ denote the tensor connection product which contracts several tensors as follows:
\begin{equation}
\begin{split}
& \left(\overline{\otimes}^{B}_{d=A}\mathcal{U}_d\right)\left(r_A,:,r_{B+1}\right) \\
=& \sum^{R_{A+1}}_{r_{A+1}=1}\dotsm \sum^{R_B}_{r_B=1}\mathcal{U}_A\left(r_A,:,r_{A+1}\right) \otimes \dotsm \otimes \\
& \mathcal{U}_B\left(r_B,:, r_{B+1}\right)
\end{split}
\end{equation}
\end{definition}

\begin{definition}[General tensor ring unfolding]
Let $\mathbf{X}_{\left\{k,L\right\}}\in \mathbb{R}^{\prod^{k+L-1}_{i=k}n_i\times \prod^{k-1}_{j=k+L}n_j}$ denotes the $k$-shifting $L$-matricization of $\mathcal{X}$ which first permutes $\mathcal{X}$ with order $\left[k,\dotsc,D,1,\dotsc,k-1\right]$ and performs matricization along first $L$ modes. The indices of $\left(\mathbf{X}_{\left\{k,L\right\}}\right)_{st}$ are formulated as
\begin{align}
& s=1+\sum_{i=k}^{k+L-1}\left( N_i-1 \right)\prod_{j=k}^{i-1}N_j \\
& t=1+\sum_{i=k+L}^{k-1}\left( N_i-1 \right)\prod_{j=k+L}^{i-1}N_j
\end{align}
\end{definition}

\section{Robust Tensor Ring Completion}
\label{section-method}

As it is proved in \cite{huang2020provable}, the rank of TR unfolding $\mathbf{X}_{\left\{d,L\right\}}$ obeys $\operatorname{rank}\left(\mathbf{X}_{\left\{d,L\right\}}\right)=R_dR_{d+L}$ under the canonical form of TR decomposition, provided that the TR rank is $\left[R_1,\dotsc,R_D\right]$. Motivated by that $\lceil D/2 \rceil$ unfoldings are sufficient for low rank tensor ring completion \cite{huang2020provable}, we formulate the corresponding optimization model for robust low rank TR completion as follows:
\begin{equation}
\label{model2-1}
\begin{split}
\min_{\mathcal{L},\; \mathcal{S}} \;& \sum_{d=1}^{\lceil D/2 \rceil}w_d{\lVert \mathbf{L}_{\left\{d, L\right\}} \rVert}_*+\lambda_d{\lVert \mathcal{S} \rVert}_1 \\
\mathrm{s.\;t.}\; &\mathscr{A}_{\mathbb{O}}\left(\mathcal{L}+\mathcal{S}\right)=\mathscr{A}_{\mathbb{O}}\left(\mathcal{T}\right),
\end{split}
\end{equation}
where $w_d$ are the weights with $\sum^{\lceil D/2 \rceil}_{d=1}w_d=1$, $\lambda_d$ are deterministic parameters, i.e. $\lambda_d=1/\sqrt{P\overline{N}_{dL}}$, where $\overline{N}_{dL}\triangleq \max\left\{\prod^{d+L-1}_{i=d}N_i,\prod^{d-1}_{i=d+L}N_i\right\}$ and the operator $\mathscr{A}_{\mathbb{O}}:\mathbb{R}^{N_1\times \dotsc \times N_D}\mapsto\mathbb{R}^{M}$ denotes the sampling process, where $M$ is the number of samples. 

The motivation for the usage of TR decomposition partly stems from the superiority in practical data completion which have been conducted in \cite{bengua2017efficient, wang2017efficient, huang2020provable}, compared with using other decompositions. Another explanation comes from the quantum mechanics, since it mention that the quantum inspired methods can capture more information  \cite{cichocki2017tensor}, though it is intractable to prove which decomposition is more beneficial in (robust) tensor completion theoretically.

As discussed in \cite{mu2014square, oymak2015simultaneously}, unlike the optimization for a single nuclear norm, the sum of the nuclear norm model is suboptimal in terms of sample complexity.  Noting that this result is for the Tucker decomposition but not the TR decomposition. Since it provides the recovery guarantee for TR completion in \cite{huang2020provable}, the near optimality of the convex surrogate model naturally follows.

\subsection{Recovery Guarantee}
\label{subsection-sampling}
 
\begin{assumption}[Strong TR incoherence assumption \cite{huang2020provable}]
\label{TRsip}
A $D$-order tensor $\mathcal{T}\in \mathbb{R}^{N_1\times \dotsm \times N_D}$ obeys the TR strong incoherence property with parameter $\boldsymbol{\mu}=\left[\mu_1,\dotsc,\mu_D\right]$, $\boldsymbol{\mu} \succ \mathbf{0}$ if for any $d \in \left\{1,\dotsc,D\right\}$,
\begin{align}
\label{TR strong incoherence property}
|\langle \mathcal{U}_d\left(:,n_d,:\right),\mathcal{U}_d\left(:,n'_d,:\right) \rangle -\frac{R_dR_{d+1}}{N_d}1_{n_d=n'_d}|\leq \notag \\
\frac{\mu_d\sqrt{R_dR_{d+1}}}{N_d},
\end{align}
where $1_{n_d=n'_d}$ takes value $1$ only if $n_d=n'_d$ and equals $0$ otherwise.
\end{assumption}

Assuming that a $D$-order tensor $\mathcal{T}\in \mathbb{R}^{N_1\times \dotsm \times N_D}$ is sampled from a uniformly bounded model with TR rank being $\left[R_1,\dotsc,R_d\right]$. Define $\overline{N}_{dL}$ and $\underline{N}_{dL}$ as the maximum and minimum values of $\left\{\prod^{d+L-1}_{i=d}N_i,\prod^{d-1}_{i=d+L}N_i\right\}$. The following condition characterizes our main result.
\begin{corollary}[Robust TR completion]
\label{corollary-Robust TR sampling bound}
Assume $\mathcal{T}$ obeys Assumption \ref{TRsip}, support set $\mathbb{O}$ is uniformly and randomly chosen from $\left\{1,\dotsc,|\mathcal{X}|\right\}$ with cardinality $M=P|\mathcal{X}|$ and each element in $\mathbb{O}$ is independently and identically corrupted with probability $\gamma$,  (\ref{model2-1})  has exact and unique solutions for $\mathcal{L}$ and $\mathcal{S}$ with probability at least $1-C\overline{N}_{dL}^{-3}$, provided that
\begin{equation}
\label{bound2}
R_dR_{d+L}\leq C_PP\underline{N}_{dL} \mu^{-1}\left(\lceil D/2 \rceil\ln\overline{N}_{dL}\right)^{-2},\;\gamma\leq C_{\gamma},
\end{equation}
where $C$, $C_P$ and $C_{\gamma}$ are positive constants, and $\mu$ can be found in Lemma 1 in \cite{huang2020provable}.
\end{corollary}
\begin{proof*}
The proof directly follows from \cite{huang2014provable} and \cite{huang2020provable} and hence is omitted.

$\hfill \blacksquare$
\end{proof*}
The above result shows that robust tensor ring completion (RTRC) simultaneously recoveries the low-rank $\mathcal{L}$ produced by TR representation and arbitrary sparse $\mathcal{S}$ with high probability, supposing each entry is observed with probability $P$ and the TR unfolding rank increases in a asymptotic linear form, specifically, on the order of $\underline{N}_{dL}\ln^{-5/2} \overline{N}_{dL}$ by replacing $\mu$ with $O\left(\ln^{1/2} \overline{N}_{dL}\right)$ \cite{huang2020provable}. It only makes assumptions about the incoherence of $\mathcal{L}$, and the magnitude and location of $\mathcal{S}$ can be arbitrary. When $D=2$, this model is consistent with the MRC model, in which $R_dR_{d+L}$ can be regarded as the matrix rank.

This model has only two kinds of parameters to tune since the parameters $\lambda_d$'s are determined by the mathematical deduction. The supremum of the sampling lower bound mainly depends on $\overline{N}^2_{dL}\ln^{5/2} \overline{N}_{dL}$, the value of $L$ can be set in a (sub)optimal way, say, fixing $L=\lceil D/2 \rceil$ for a tensor with comparable size. Another perspective to comprehend this selection is Von Neumann entropy \cite{bengtsson2017geometry}, which is defined as $S=-\operatorname{tr}\left(\rho\ln \rho\right)$ where $\rho$ represents the density matrix. We can use this entropy to characterize the correlations between different tensor dimensions. From its strong subadditivity, i.e., for any three systems $A$, $B$ and $C$ we have  $S\left(\rho_{ABC}\right)+S(\rho_B)\leq S\left(\rho_{AB}\right)+S\left(\rho_{BC}\right)$, it shows that a balanced entanglement of different systems has greater entropy, which means more information can be captured if we unfold the TR in a balanced way and this is critical to promote the completion performance, especially for practical data.

Since the weights only affect the constants in Corollary \ref{corollary-Robust TR sampling bound}, the weights $w_d$'s are set by normalizing a vector consisting of the reciprocals of the supremums $\overline{N}^2_{dL}\ln^{5/2} \overline{N}_{dL}$ accordingly. The patterns of practical data often do not match the TR structure perfectly and hence the recovery performance deteriorates compared with synthetic data. By the aforementioned weight setting the algorithm is forced to recover the unfoldings that have lower sample complexity. In fact, the practical performance relies on the chosen rank-revealing structure, albeit the selection is intractable theoretically and beyond the scope of this paper.

Her we use more than one unfolding in order to obtain better performance in practical data completion, and capture the correlations between different tensor modes as much as possible.

\subsection{Algorithm}
\label{subsection-RTRC}

The variable $\mathbf{X}^{\left(d\right)}$ to is used to replace $\mathbf{L}_{\left\{d,L\right\}}$ in  (\ref{model2-1}), which yields the following model:
\begin{equation}
\label{model2-2}
\begin{split}
\min_{\mathcal{X}^{\left(d\right)},\;\mathcal{L},\;\mathcal{S}} \;& \sum_{d=1}^{\lceil D/2 \rceil}w_d{\lVert \mathcal{X}^{\left(d\right)}_{\left\{ d,L \right\}} \rVert}_*+\lambda_d{\lVert \mathcal{S} \rVert}_1 \\
\mathrm{s.\;t.}\;\;\; &\mathscr{A}_{\mathbb{O}}\left(\mathcal{L}+\mathcal{S}\right)=\mathscr{A}_{\mathbb{O}}\left(\mathcal{T}\right), \\
& \mathcal{X}^{\left(d\right)}=\mathcal{L}\;\;\left(d=1,\dotsc,\lceil D/2 \rceil\right).
\end{split}
\end{equation}
Its augmented Lagrangian function is
\begin{equation}
\label{model2-3}
\begin{split}
& \mathscr{L}_{\beta}\left(\mathcal{X}^{\left(d\right)},\mathcal{L},\mathcal{S}\right)=\lambda{\lVert \mathcal{S} \rVert}_1+\sum_{d=1}^{\lceil D/2 \rceil}w_d{\lVert \mathcal{X}^{\left(d\right)}_{\left\{ d,L \right\}} \rVert}_*+ \\
& \langle \mathcal{Z}^{\left(d\right)},\mathcal{X}^{\left(d\right)}-\mathcal{L} \rangle+\frac{\beta_d}{2}\lVert \mathcal{X}^{\left(d\right)}-\mathcal{L} \rVert^2_{\mathrm{F}}+ \\
& \langle \mathbf{w},\mathscr{A}_{\mathbb{O}}\left(\mathcal{L+S-T}\right) \rangle+\frac{\beta}{2}\lVert \mathscr{A}_{\mathbb{O}}\left(\mathcal{L+S}\right)-\mathscr{A}_{\mathbb{O}}\left(\mathcal{T}\right) \rVert^2_2,
\end{split}
\end{equation}
where $\lambda=\sum^{\lceil D/2 \rceil}_{d}\lambda_d$, $\beta_d$ and $\beta$ are penalty coefficients, and $\mathcal{Z}^{\left(d\right)}$ and $\mathbf{w}$ are dual variables. ADMM can divide the problem into $5$ sub-problems as follows.

\subsubsection{Update of $\mathcal{X}$}

Considering $\mathscr{L}_{\beta}$ as a function of $\mathcal{X}^{\left(d\right)}$ while $\mathcal{L}$ and $\mathcal{S}$ are fixed, we have an equivalent problem:
\begin{equation*}
\mathcal{X}^{\left(d\right)*}=\mathop{\arg\min}_{\mathcal{X}^{\left(d\right)}}\; \frac{w_d}{\beta_d}\lVert \mathcal{X}^{\left(d\right)}_{\left\{ d,L \right\}} \rVert_*+\frac{1}{2}\lVert \mathcal{X}^{\left(d\right)}-\left(\mathcal{L}-\frac{1}{\beta_d}\mathcal{Z}^{\left(d\right)}\right) \rVert^2_{\mathrm{F}}.
\end{equation*}
This problem has a closed-form solution: 
\begin{equation}
\label{update2-X}
\mathcal{X}^{\left(d\right)*}=\operatorname{D}_{\frac{w_d}{\beta_d}}\left(\mathcal{L}-\frac{1}{\mu_d}\mathcal{Z}^{\left(d\right)}\right),
\end{equation}
where $\operatorname{D}\left(\cdot\right)$ is the singular value thresholding operator \cite{ma2011fixed}.

\subsubsection{Update of $\mathcal{L}$}

Denote by $\mathscr{A}^*_{\mathbb{O}}$ the adjoint of $\mathscr{A}_{\mathbb{O}}$ and note that $\mathscr{A}^*_{\mathbb{O}}\left(\mathbf{w}\right)=\mathscr{P}_{\mathbb{O}}\left(\mathcal{W}\right)$ and $\mathscr{A}_{\mathbb{O}}^*\mathscr{A}_{\mathbb{O}}=\mathscr{P}_{\mathbb{O}}$, the optimality condition w.r.t. $\mathcal{L}$ is 
\begin{equation*}
\begin{split}
\left(\lceil D/2 \rceil\mathscr{E}+\mathscr{P}_{\mathbb{O}}\right)\left(\mathcal{L}\right)=& \sum^{\lceil D/2 \rceil}_{d=1}\left[\mathcal{X}^{\left(d\right)}+\frac{1}{\beta_d}\mathcal{Z}^{\left(d\right)}\right]+ \\
& \mathscr{P}_{\mathbb{O}}\left(\mathcal{T}-\mathcal{S}-\frac{1}{\mu}\mathcal{W}\right),
\end{split}
\end{equation*}
where $\mathscr{E}$ is the identity operator defined previously. Solving this linear system leads to
\begin{equation}
\label{update2-L}
\begin{split}
\mathcal{L}^*=& \left\{\sum^{\lceil D/2 \rceil}_{d=1}\left[\mathcal{X}^{\left(d\right)}+\frac{1}{\beta_d}\mathcal{Z}^{\left(d\right)}\right]+\mathcal{P}\circledast\left(\mathcal{T}-\mathcal{S}-\frac{1}{\beta}\mathcal{W}\right)\right\}\oslash \\
& \left(\lceil D/2 \rceil\mathcal{I}+\mathcal{P}\right),
\end{split}
\end{equation}
where $\mathcal{P}$ is the binary sampling tensor and $\oslash$ represents the element-wise division.

\subsubsection{Update of $\mathcal{S}$}

Similar to the update of $\mathcal{L}$, $ \mathcal{S} $ is updated by the following optimization model: 
\begin{equation*}
\min_{\mathcal{S}}\; \frac{1}{2}\lVert \mathscr{A}_{\mathbb{O}}\left(\mathcal{S}\right)-\mathscr{A}_{\mathbb{O}}\left(\mathcal{T}-\mathcal{L}-\frac{1}{\beta}\mathcal{W}\right) \rVert^2_{\mathrm{F}}+\frac{\lambda}{\beta}\lVert \mathcal{S} \rVert_1.
\end{equation*}
According to Lemma \ref{lemma1}, the optimal solution is
\begin{equation}
\label{update2-S}
\mathcal{S}^*=\operatorname{S}_{\frac{\lambda}{\beta}}\left(\mathcal{P}\circledast\left(\mathcal{T}-\mathcal{L}-\frac{1}{\beta}\mathcal{W}\right)\right),
\end{equation}
where $\operatorname{S}\left(\cdot\right)$ is the soft thresholding operator \cite{ma2011fixed}.

\begin{lemma}
\label{lemma1}
The solution to the optimization model:
\begin{equation*}
\min_{\mathcal{X}}\frac{1}{2}{\lVert \mathscr{A}_{\mathbb{O}}\left(\mathcal{X}\right)-\mathscr{A}_{\mathbb{O}}\left(\mathcal{B}\right) \rVert}^2_2+\tau\lVert \mathcal{X} \rVert_1
\end{equation*}
is $\mathcal{X}^*=\operatorname{S}_{\tau}\left(\mathcal{P}\circledast\mathcal{B}\right)$, where $\mathcal{P}$ is the binary sampling tensor.
\end{lemma}

\subsubsection{Update of $\mathcal{Z}$}

The update of the dual variable $\mathcal{Z}^{\left(d\right)}$ is given by
\begin{equation}
\label{update2-Z}
\mathcal{Z}^{\left(d\right)}=\mathcal{Z}^{\left(d\right)}+\beta_d\left(\mathcal{X}^{\left(d\right)}-\mathcal{L}\right).
\end{equation}

\subsubsection{Update of $\mathcal{W}$}

According to the rule of ADMM, the vector form of update is
\begin{equation}
\label{update2-W}
\mathcal{W}=\mathcal{W}+\beta\mathcal{P}\circledast\left(\mathcal{L+S-T}\right).
\end{equation}

The details about the proposed solution for RTRC is summarized in Algorithm \ref{algorithm-RTRC}, where the Lanczos algorithm is used for fast singular value decomposition \cite{liu2016generalized, yang2013fixed}.
\begin{algorithm}
\caption{robust tensor ring completion (RTRC) via alternating direction method of multipliers}
\label{algorithm-RTRC}
\begin{algorithmic}[1]
\REQUIRE Zero-filled observed tensor $\mathcal{T}$, observation set $\mathbb{O}$, penalty coefficient $\boldsymbol{\beta}=\left[ \beta_1,\dotsc,\beta_{\lceil D/2 \rceil},\beta \right]$, the maximal \# iterations $K$.
\ENSURE Recovered low-rank tensor $\mathcal{L}$ and sparse tensor $\mathcal{S}$.
\STATE Initialization $\mathcal{P}$, $\mathcal{L}_0=\mathcal{P\circledast T}$, $\mathcal{S}_0=\mathcal{O}$, $\left\{\mathcal{X}\right\}=\mathcal{L}_0$, $\left\{\mathcal{Z}\right\}=\mathcal{W}=\mathcal{O}$.
\FOR{$k=1\;\TO\;K$}
\FOR{$d=1\;\TO\;\lceil D/2 \rceil$}
\STATE Update $\mathcal{X}^{\left(d\right)}$ according to (\ref{update2-X})
\ENDFOR
\STATE Update $\mathcal{L}$ according to (\ref{update2-L})
\STATE Update $\mathcal{S}$ according to (\ref{update2-S})
\FOR{$d=1\;\TO\;\lceil D/2 \rceil$}
\STATE Update $\mathcal{Z}^{\left(d\right)}$ according to (\ref{update2-Z})
\ENDFOR
\STATE Update $\mathcal{W}$ according to (\ref{update2-W})
\ENDFOR
\RETURN{$\mathcal{L}$ and $\mathcal{S}$}
\end{algorithmic}
\end{algorithm}

\subsection{Algorithmic Complexity}

For a $D$-order tensor $\mathcal{X}\in \mathbb{R}^{N\times \dotsm \times N}$ with TR rank $\left[R,\dotsc,R\right]$. Since Lanczos method has a linear complexity $O\left(I_1+I_2\right)$ for a $I_1$-by-$I_2$ matrix, the complexity of RTRC algorithm mainly depend on the updates of $\mathcal{X}$ which involve $D/2$ soft thresholdings and hence cost $O\left(DN^{D/2}\right)$.

The storage complexity is $DN^{D/2}R^2$ because $D/2$ outcomes of SVDs are stored.

\subsection{Algorithmic Convergence}

The ADMM has a linear rate of convergence when one of the objective terms is strongly convex \cite{nishihara2015general}. To check this we can refer to the $\ell_1$ term in regularization. Reference \cite{lin2010augmented} provides a rather simple but efficient strategy to improve convergence. They suggest that the penalty coefficient increases geometrically with iterations. In details, the $\beta_d$'s and $\beta$ in (\ref{model2-3}) follow $\beta^{k+1}_d=\alpha\mu^k_d$ and $\beta^{k+1}=\alpha\mu^k$, where $\alpha$ is some numerical constant.

\section{Numerical Experiments}
\label{section-experiment}

In this section, four groups of datasets are used for tensor completion experiments, i.e., synthetic data, face images, light-filed images and color videos. 

Eight algorithms are used to test the performance on real-world data, i.e., robust tensor completion via tensor nuclear norm minimization (RTC-TNN) based on t-SVD \cite{wang2019robust}, robust tensor completion via sum of matrix nuclear norm minimization (RTC-SNN) based Tucker decomposition \cite{goldfarb2014robust}, robust matrix completion (RMC) \cite{candes2011robust}, simple low rank tensor completion via tensor train decomposition (SiLRTC-TT) \cite{bengua2017efficient}, low rank tensor tree decomposition for tensor completion (STTC) \cite{liu2019image}, tensor ring nuclear norm minimization for tensor completion (TRNNM) \cite{yu2019tensor}, Bayesian CANDECOMP/PARAFAC factorization (FBCP) for image recovery \cite{zhao2015bayesian} and the proposed RTRC.

All the experiments are conducted by MATLAB 9.3.0 on a desktop with a 2.8GHz CPU of Intel Core i7 and a 16GB RAM.

Three performance evaluation metrics are used. Relative error (RE) is defined as $\text{RE}=\lVert \hat{\mathcal{X}}-\mathcal{X} \rVert_{\mathrm{F}}/{\lVert \mathcal{X} \rVert}_{\mathrm{F}}$, where $\mathcal{X}$ is the ground truth and $\hat{\mathcal{X}}$ is the recovered tensor. The second one termed peak signal-to-noise ratio (PSNR) is the ratio between the maximum possible power of a signal and the power of corrupting noise \cite{barnsley1993fractal}. The third one evaluates  algorithmic complexity in terms of computational CPU time.

As defined previously, the sampling ratio $P$ represents the ratio of the number of sampled entries to the cardinality of tensor $\mathcal{X}$, which is denoted as $P= \left| \mathbb{O} \right| / \left| \mathcal{X} \right|  $.

For fair comparisons, the parameters in each algorithm are tuned to give the optimal performance. The convergence is judged by the relative change (RC) \text{RC}=$\lVert \mathcal{L}^k-\mathcal{L}^{k-1}\rVert_{\mathrm{F}}/\lVert \mathcal{L}^{k-1}\rVert_{\mathrm{F}}$, where the tolerance parameter is set to be $1\times 10^{-8}$. The number of maximal iterations is $200$. The penalty coefficients are set as $\beta_1=\dotsm=\beta_{\lceil D/2 \rceil}=\beta$.

\subsection{Exact Recovery of Synthetic Data}
\label{subsection-exact recovery}

We simulate a low rank tensor $\mathcal{L} \in \mathbb{R}^{N\times N \times N \times N}$ by TR contraction, with $ N = 10 $ or $ N = 20$. The entries of each factor are i.i.d. standard Gaussian random variables, i.e., $\mathcal{U}_d\left(r_d,n_d,r_{d+1}\right) \sim \mathcal{N}\left(0,1/N_d\right)$, $d=1,\dotsc,D$. The observation location $\mathbb{O}$ is randomly chosen from $\left\{1,\dotsc,N^D\right\}$,  and the sparse noise $\mathcal{S}_0$ is generated by $\mathscr{A}_{\mathbb{O}}\left(\mathcal{S}_0\right)=\mathbf{s}$ and $\mathscr{A}_{ \mathbb{O}^\perp}\left(\mathcal{S}_0\right)=\mathbf{0}$, where $\mathbf{s}$ is a Bernoulli vector whose entries is uniformly distributed on $\left\{\pm1\right\}$. In this experiment, we set $\alpha=1.1$ and $\beta^0=1\times10^{-2}$.

We test the algorithm's recovery ability on two conditions: $P=1$ and $0.81$. Each condition contains four subgroups: $\operatorname{rank}\left(\mathcal{L}_0\right)=0.2N$ and $0.3N$, $\lVert \hat{\mathcal{S}} \rVert_0=0.05M$ and $0.1M$. For each parameter setting we repeatedly perform the experiment $10$ times. 

Table. \ref{result-synthetic data 1} reports the averaged recovery result of randomly generated problems. The result shows that the RTRC correctly solves these problems, where $\lVert \hat{\mathcal{L}}-\mathcal{L}_0 \rVert_{\mathrm{F}}/\lVert \mathcal{L}_0 \rVert_{\mathrm{F}}< 10^{-6}$ is considered to be successfully recovered. It can be seen that most estimations of TR rank and sparsity are correct and the relative error $\lVert \hat{\mathcal{S}}-\mathcal{S}_0 \rVert_{\mathrm{F}}/\lVert \mathcal{S}_0 \rVert_{\mathrm{F}}<10^{-5}$. As we can see, the TR rank with $P=0.81$ is $0.9$ times as much as that with $P=1$, which well verifies the recovery guarantee of Corollary \ref{corollary-Robust TR sampling bound}. The reason is that Corollary \ref{corollary-Robust TR sampling bound} shows that when $P$ drops from $1$ to $0.81$, the acceptable TR rank changes from $R$ to $\sqrt{P}R$.

\begin{table*}
\centering
\caption{Correct recovery result of various randomly generated problems.}
\label{result-synthetic data 1}
\begin{tabular}{c|c|c|c|c|c|c|c}
\toprule
Size $N$ & \#samples $M$ & $\operatorname{rank}\left(\mathcal{L}_0\right)$ & $\lVert \mathcal{S}_0 \rVert_0$ & $\operatorname{rank}\left(\hat{\mathcal{L}}\right)$ & $\lVert \hat{\mathcal{S}} \rVert_0$ & $\frac{\lVert \hat{\mathcal{L}}-\mathcal{L}_0 \rVert_{\mathrm{F}}}{\lVert \mathcal{L}_0 \rVert_{\mathrm{F}}}$ & $\frac{\lVert \hat{\mathcal{S}}-\mathcal{S}_0 \rVert_{\mathrm{F}}}{\lVert \mathcal{S}_0 \rVert_{\mathrm{F}}}$ \\
\hline \hline
\multirow{2}{*}{10} & 10000 & 2 & 500 & 2 & 500 & $1.24\times 10^{-7}$ & $1.17\times 10^{-6}$ \\
\cline{2-8}
& 8100 & 2 & 405 & 2 & 405 & $2.02\times 10^{-7}$ & $1.66\times 10^{-6}$ \\
\hline
\multirow{2}{*}{20} & 160000 & 4 & 8000 & 4 & 8000 & $2.34\times 10^{-8}$ & $1.23\times 10^{-6}$ \\
\cline{2-8}
& 129600 & 3 & 6480 & 3 & 6480 & $7.76\times 10^{-8}$ & $1.49\times 10^{-6}$ \\
\hline
\multicolumn{8}{c}{$\operatorname{rank}\left(\mathcal{L}_0\right)=0.2N,\; \lVert \mathcal{S}_0 \rVert_0=0.05M\;(M=N^4\;and\;0.81N^4)$} \\
\multicolumn{8}{c}{} \\

\hline
\multirow{2}{*}{10} & 10000 & 3 & 500 & 3 & 500 & $1.58\times 10^{-7}$ & $7.37\times 10^{-6}$ \\
\cline{2-8}
& 8100 & 2 & 405 & 2 & 405 & $4.28\times 10^{-7}$ & $4.46\times 10^{-6}$ \\
\hline
\multirow{2}{*}{20} & 160000 & 6 & 8000 & 6 & 8000 & $1.16\times 10^{-8}$ & $1.36\times 10^{-6}$ \\
\cline{2-8}
& 129600 & 5 & 6480 & 5 & 6480 & $3.53\times 10^{-8}$ & $1.66\times 10^{-6}$ \\
\hline
\multicolumn{8}{c}{$\operatorname{rank}\left(\mathcal{L}_0\right)=0.3N,\; \lVert \mathcal{S}_0 \rVert_0=0.05M\;(M=N^4\;and\;0.81N^4)$} \\
\multicolumn{8}{c}{} \\

\hline
\multirow{2}{*}{10} & 10000 & 2 & 1000 & 2 & 1000 & $2.21\times 10^{-7}$ & $1.42\times 10^{-6}$ \\
\cline{2-8}
& 8100 & 2 & 810 & 2 & 810 & $1.47\times 10^{-7}$ & $1.46\times 10^{-6}$ \\
\hline
\multirow{2}{*}{20} & 160000 & 4 & 16000 & 4 & 16000 & $4.69\times 10^{-8}$ & $1.42\times 10^{-6}$ \\
\cline{2-8}
& 129600 & 3 & 12960 & 3 & 12960 & $1.13\times 10^{-7}$ & $1.61\times 10^{-6}$ \\
\hline
\multicolumn{8}{c}{$\operatorname{rank}\left(\mathcal{L}_0\right)=0.2N,\; \lVert \mathcal{S}_0 \rVert_0=0.1M\;(M=N^4\;and\;0.81N^4)$} \\
\multicolumn{8}{c}{} \\

\hline
\multirow{2}{*}{10} & 10000 & 3 & 1000 & 3 & 1000 & $1.11\times 10^{-7}$ & $2.48\times 10^{-6}$ \\
\cline{2-8}
& 8100 & 2 & 810 & 2 & 810 & $4.42\times 10^{-7}$ & $4.34\times 10^{-6}$ \\
\hline
\multirow{2}{*}{20} & 160000 & 6 & 16000 & 6 & 16000 & $1.42\times 10^{-8}$ & $1.25\times 10^{-6}$ \\
\cline{2-8}
& 129600 & 5 & 12960 & 5 & 12932 & $6.20\times 10^{-3}$ & $3.93\times 10^{-1}$ \\
\hline
\multicolumn{8}{c}{$\operatorname{rank}\left(\mathcal{L}_0\right)=0.3N,\; \lVert \mathcal{S}_0 \rVert_0=0.1M\;(M=N^4\;and\;0.81N^4)$} \\
\bottomrule
\end{tabular}
\end{table*}

\subsection{Phase Transition in TR rank and Sparsity with Varying Sampling Ratios}
\label{subsection-phase transition}

\begin{figure}[htbp]
\centering
\includegraphics[scale=0.47]{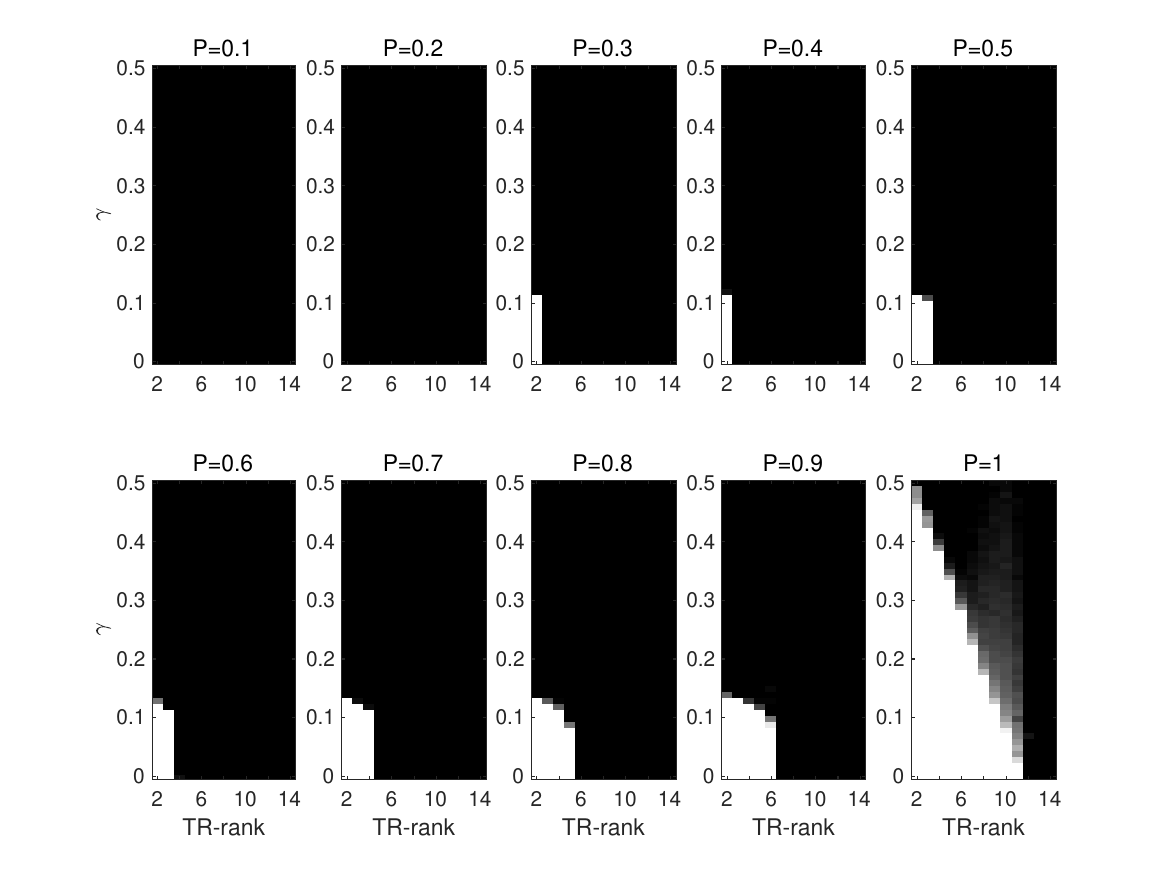}
\caption{Average recovery result over $10$ trials of a $20\times 20\times 20\times 20$ tensor with varying TR ranks, sparsity degrees and sampling ratios.}
\label{result-phase transition}
\end{figure}

To verify Corollary \ref{corollary-Robust TR sampling bound}, we consider a $4$-order tensor with $N_1=N_2=N_3=N_4=20$. The TR rank varies from $2$ to $14$. The sparsity degree $\gamma$ changes from $0$ to $0.5$ with linear increment $0.1$. The sampling ratio $P$ ranges from $0.1$ to $0.9$ with linear increment $0.1$. For each condition $\left(\operatorname{rank}\left(\mathcal{L}_0\right),\lVert \mathcal{S}_0 \rVert_0,P\right)$, we run the RTRC algorithm $10$ times to get the averaged recovery. In this experiment, we set $\alpha=1.1$ and $\beta^0=1\times10^{-2}$. The phase transition is shown in Fig. \ref{result-phase transition}, in which a black patch means a failure and a white patch means a success ($\lVert \hat{\mathcal{L}}-\mathcal{L}_0 \rVert_{\mathrm{F}}/\lVert \mathcal{L}_0 \rVert_{\mathrm{F}}< 10^{-3}$). It can be concluded that the exact recovery in accordance with the result in Corollary \ref{corollary-Robust TR sampling bound}, i.e., the TR rank is on the order of $PN^2/\left(\mu\ln^2\left(N^2\right)\right)$ and the sparsity on is the order of $N^4$. The result shows that when $P=1$ the recovery result is similar to that in \cite{candes2011robust}, and with $P$ decreasing the ``successful area'' gradually shrinks to a small piece at  rate $\sqrt{P}$, which is a validation of Corollary \ref{corollary-Robust TR sampling bound}.

\subsection{Light Field Images: Application to Image Recovery}
\label{subsection-light filed image}

\begin{figure}[htbp]
\centering
\begin{subfigure}[t]{0.11\textwidth}
\centering
\setlength{\abovecaptionskip}{0pt}
\setlength{\belowcaptionskip}{-2pt}
\includegraphics[scale=0.8]{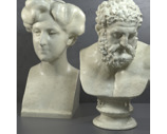}
\subcaption*{\emph{Greek}}
\end{subfigure}
\begin{subfigure}[t]{0.11\textwidth}
\centering
\setlength{\abovecaptionskip}{0pt}
\setlength{\belowcaptionskip}{-2pt}
\includegraphics[scale=0.8]{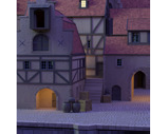}
\subcaption*{\emph{Medieval2}}
\end{subfigure}
\begin{subfigure}[t]{0.11\textwidth}
\centering
\setlength{\abovecaptionskip}{0pt}
\setlength{\belowcaptionskip}{-2pt}
\includegraphics[scale=0.8]{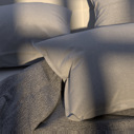}
\subcaption*{\emph{Pillows}}
\end{subfigure}
\begin{subfigure}[t]{0.11\textwidth}
\centering
\setlength{\abovecaptionskip}{0pt}
\setlength{\belowcaptionskip}{-2pt}
\includegraphics[scale=0.8]{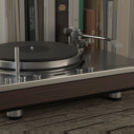}
\subcaption*{\emph{Vinyl}}
\end{subfigure}
\caption{The centre view of four $4$-D light field images.}
\label{image dataset 2}
\end{figure}

The dataset used in this subsection contains four  $4$-D light field images: \emph{greek}, \emph{medieval2}, \emph{pillows} and \emph{vinyl} \footnote{http://hci-lightfield.iwr.uni-heidelberg.de}. They are all down-sampled to be with the size of $128\times128\times3\times81$, as it shows in Fig. \ref{image dataset 2}. 

For each image, we randomly choose $30\%$ pixels as observation and randomly choose $10\%$ entries from the observation with their values being randomly distributed in $\left[0,255\right]$. The frames are of low-rank since they are similar. The parameters of the compared algorithms are set as suggested in \cite{goldfarb2014robust, wang2019robust}. The parameter settings of our method are as follows: $\alpha=1.1$ and $\beta^0=1\times10^{-4}$. We repeat the experiment with respect to each image $10$ times for avoiding coincidence.

\begin{figure}[htbp]
\centering
\includegraphics[scale=0.22]{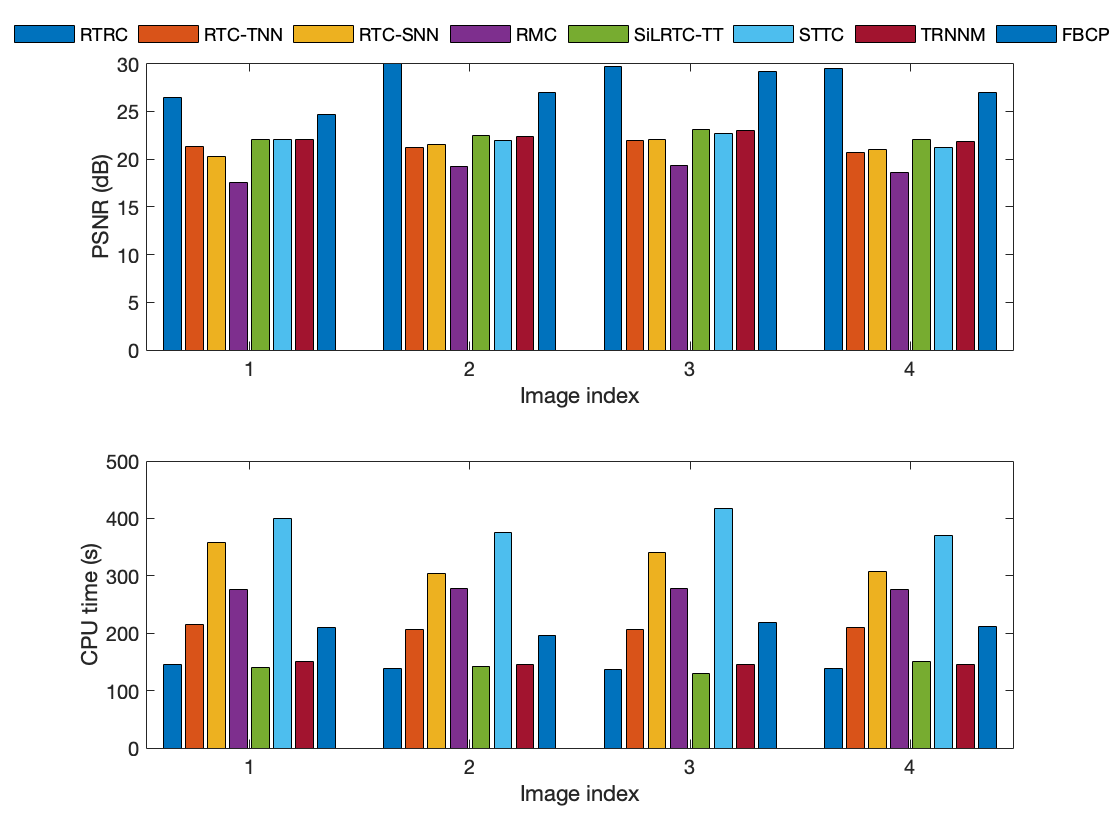}
\caption{ The recovery results of four light-field images performed by eight algorithms, including RTRC, RTC-TNN, RTC-SNN, RMC, SiLRTC-TT, STTC, TRNNM and FBCP.}
\label{result-lfimage}
\end{figure}

Fig. \ref{result-lfimage} provides the recovery performance in terms of  PSNR and CPU time of four compared algorithms at an average of $10$ experiments for each image.  All the tensor-based methods show superior performance than matrix-based one, which confirms that tensor-based methods can better exploit data structure for high-order data than matrix-based methods. In addition, the proposed method outperforms the others in terms of both PSNR and CPU time.

\subsection{YaleB Face Images: Application to Shadow Removal}
\label{subsection-YaleB face image}

\begin{figure}[htbp]
\centering
\begin{subfigure}[t]{0.5\textwidth}
\centering
\setlength{\abovecaptionskip}{0pt}
\setlength{\belowcaptionskip}{-2pt}
\includegraphics[scale=0.35]{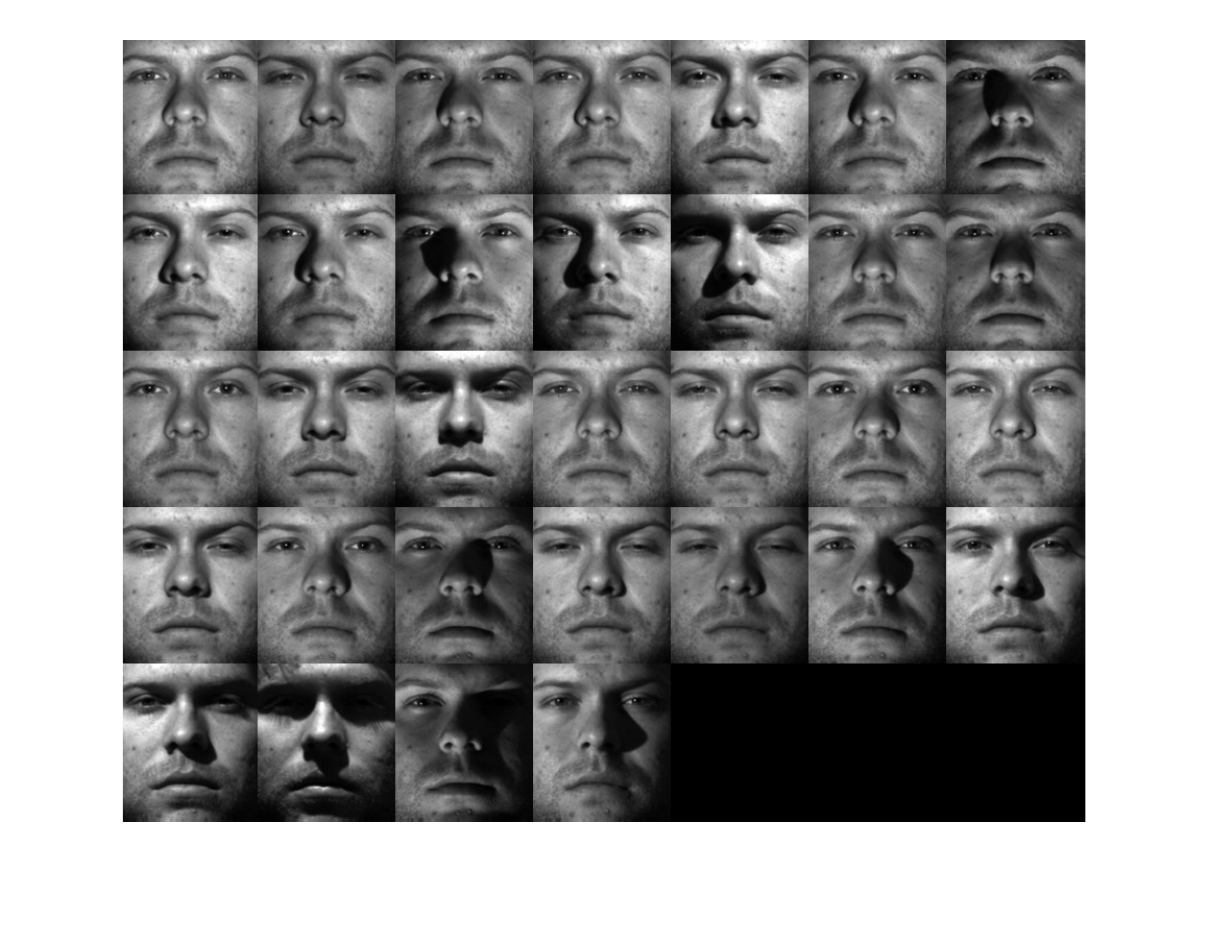}
\subcaption*{Subject 1}
\end{subfigure}

\begin{subfigure}[t]{0.5\textwidth}
\centering
\setlength{\abovecaptionskip}{0pt}
\setlength{\belowcaptionskip}{-2pt}
\includegraphics[scale=0.35]{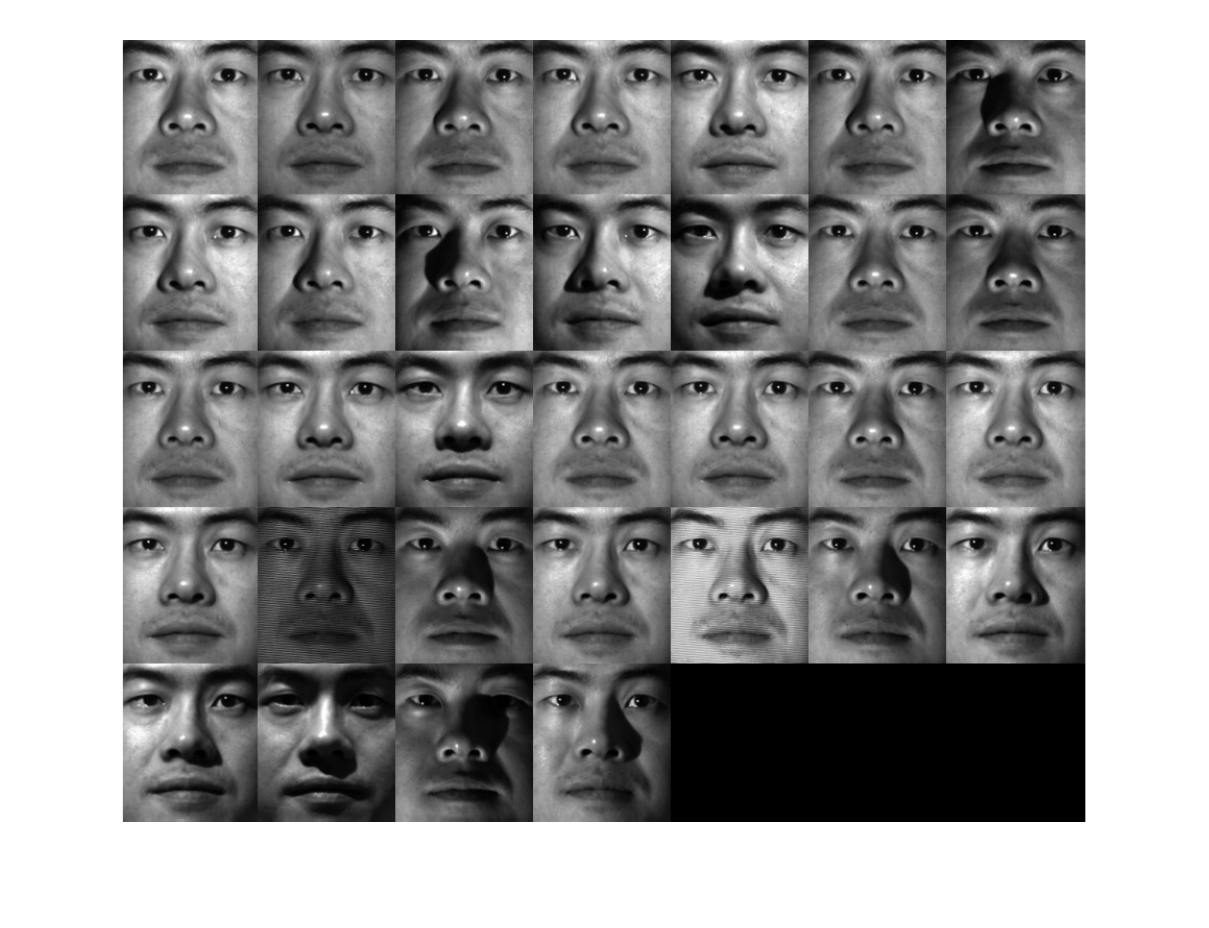}
\subcaption*{Subject 2}
\end{subfigure}
\caption{Two sequences of images from YaleB face dataset.}
\label{image dataset 3}
\end{figure}

The YaleB face dataset\footnote{http://vision.ucsd.edu/content/yale-face-database} contains $16128$ images of $28$ human subjects under $9$ poses and $64$ illumination conditions. We choose the $1$st and $2$nd subjects and extract their $1$st to $12$th, $30$th to $34$th and $36$th to $50$th frames as the dataset in this subsection, as it is shown in Fig. \ref{image dataset 3}. The latent tensor that consists of clear faces is low-rank and the various illuminations are sparse components. Removing the shadow of faces is more challenging since the shadow's locations are not as uniform as those in light field images. For each subject we randomly choose $50\%$ pixels from fully observations. We stack each frame as column in matrix in order to use RMC. We set $\alpha=1.1$ and $\beta^0=1\times 10^{-4}$ in our algorithm. We repeat the experiment with respect to each image $10$ times for avoiding coincidence.

\begin{figure*}[htbp]
\centering
\begin{subfigure}[t]{0.09\textwidth}
\centering
\setlength{\abovecaptionskip}{0pt}
\setlength{\belowcaptionskip}{-2pt}
\includegraphics[scale=0.35]{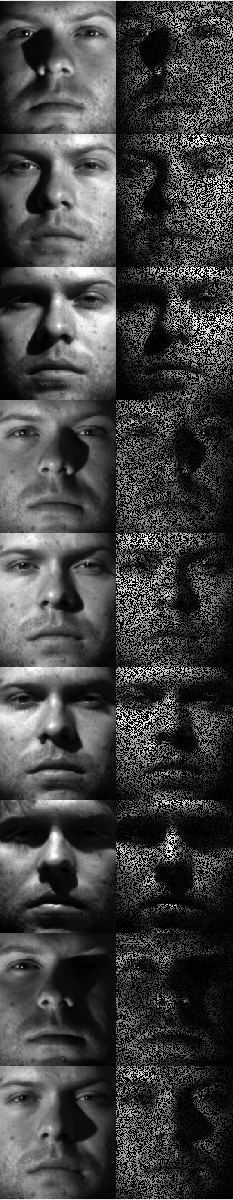}
\subcaption*{Subject 1, observation}
\end{subfigure}
\begin{subfigure}[t]{0.09\textwidth}
\centering
\setlength{\abovecaptionskip}{0pt}
\setlength{\belowcaptionskip}{-2pt}
\includegraphics[scale=0.35]{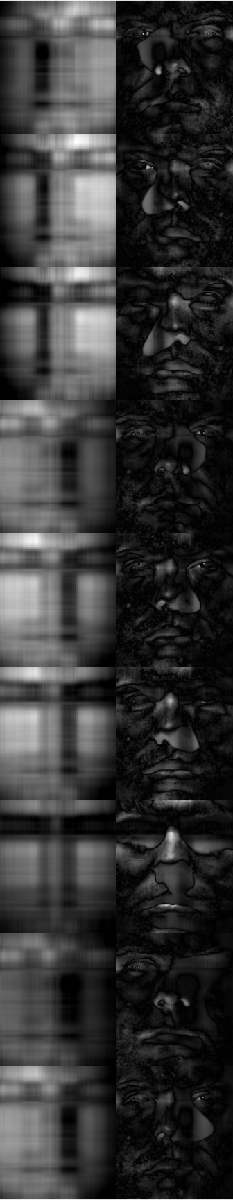}
\subcaption*{RTRC}
\end{subfigure}
\begin{subfigure}[t]{0.09\textwidth}
\centering
\setlength{\abovecaptionskip}{0pt}
\setlength{\belowcaptionskip}{-2pt}
\includegraphics[scale=0.35]{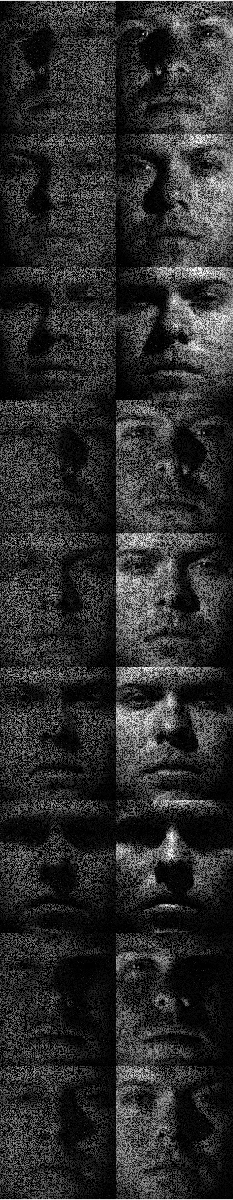}
\subcaption*{RTC-TNN}
\end{subfigure}
\begin{subfigure}[t]{0.09\textwidth}
\centering
\setlength{\abovecaptionskip}{0pt}
\setlength{\belowcaptionskip}{-2pt}
\includegraphics[scale=0.35]{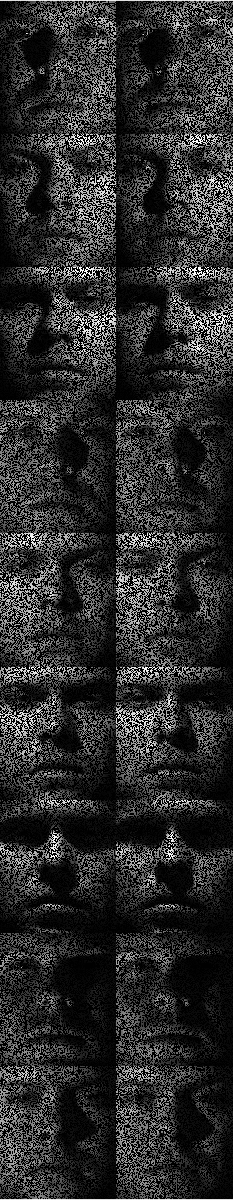}
\subcaption*{RTC-SNN}
\end{subfigure}
\begin{subfigure}[t]{0.09\textwidth}
\centering
\setlength{\abovecaptionskip}{0pt}
\setlength{\belowcaptionskip}{-2pt}
\includegraphics[scale=0.35]{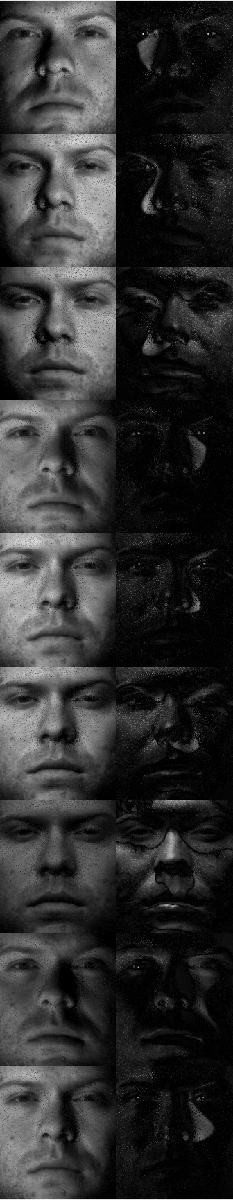}
\subcaption*{RMC}
\end{subfigure}
\begin{subfigure}[t]{0.09\textwidth}
\centering
\setlength{\abovecaptionskip}{0pt}
\setlength{\belowcaptionskip}{-2pt}
\includegraphics[scale=0.35]{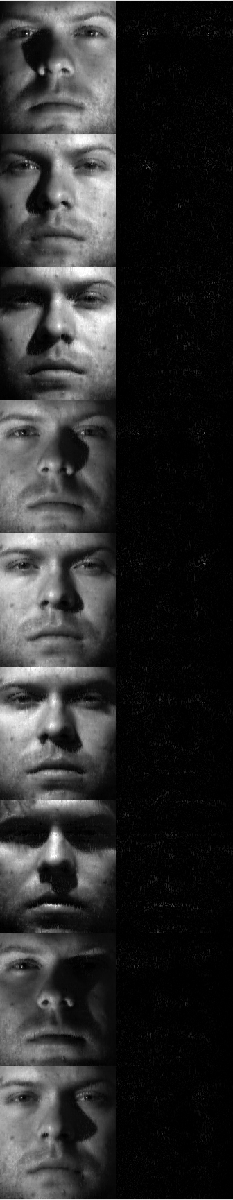}
\subcaption*{SiLRTC-TT}
\end{subfigure}
\begin{subfigure}[t]{0.09\textwidth}
\centering
\setlength{\abovecaptionskip}{0pt}
\setlength{\belowcaptionskip}{-2pt}
\includegraphics[scale=0.35]{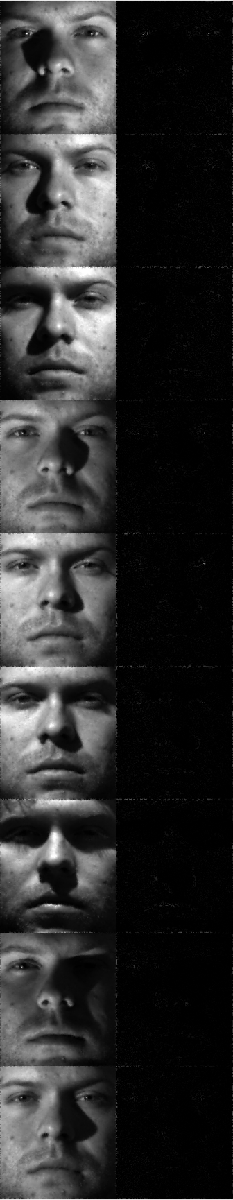}
\subcaption*{STTC}
\end{subfigure}
\begin{subfigure}[t]{0.09\textwidth}
\centering
\setlength{\abovecaptionskip}{0pt}
\setlength{\belowcaptionskip}{-2pt}
\includegraphics[scale=0.35]{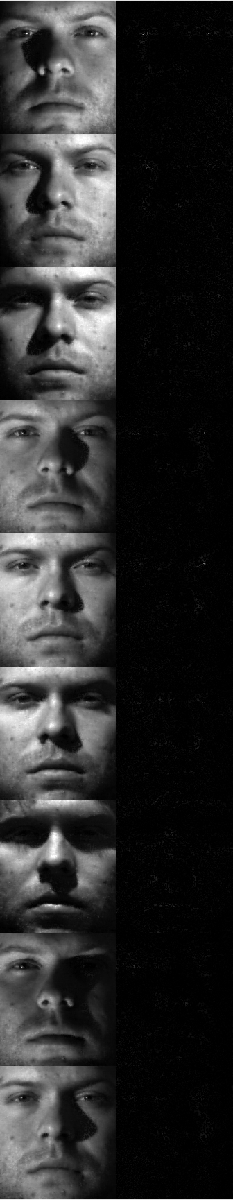}
\subcaption*{TRNNM}
\end{subfigure}
\begin{subfigure}[t]{0.09\textwidth}
\centering
\setlength{\abovecaptionskip}{0pt}
\setlength{\belowcaptionskip}{-2pt}
\includegraphics[scale=0.35]{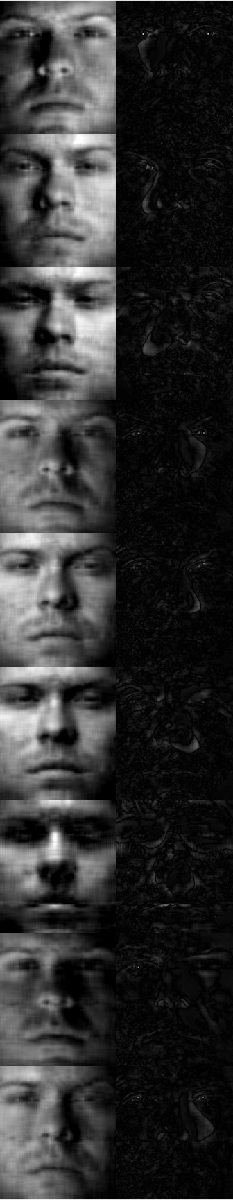}
\subcaption*{FBCP}
\end{subfigure}

\begin{subfigure}[t]{0.09\textwidth}
\centering
\setlength{\abovecaptionskip}{0pt}
\setlength{\belowcaptionskip}{-2pt}
\includegraphics[scale=0.35]{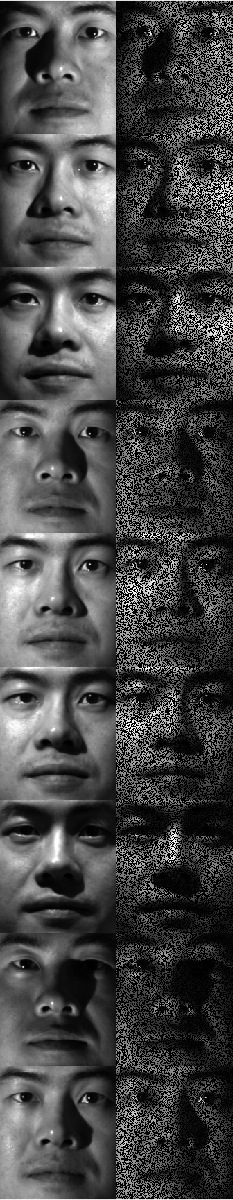}
\subcaption*{Subject 1, observation}
\end{subfigure}
\begin{subfigure}[t]{0.09\textwidth}
\centering
\setlength{\abovecaptionskip}{0pt}
\setlength{\belowcaptionskip}{-2pt}
\includegraphics[scale=0.35]{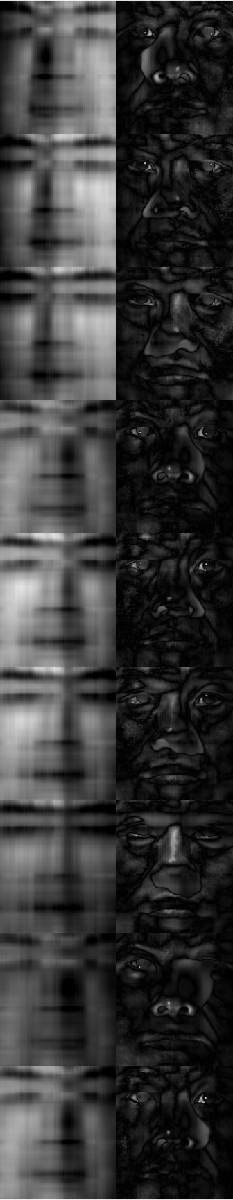}
\subcaption*{RTRC}
\end{subfigure}
\begin{subfigure}[t]{0.09\textwidth}
\centering
\setlength{\abovecaptionskip}{0pt}
\setlength{\belowcaptionskip}{-2pt}
\includegraphics[scale=0.35]{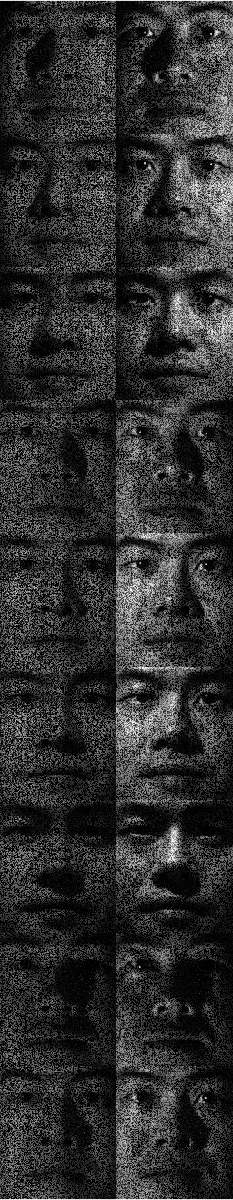}
\subcaption*{RTC-TNN}
\end{subfigure}
\begin{subfigure}[t]{0.09\textwidth}
\centering
\setlength{\abovecaptionskip}{0pt}
\setlength{\belowcaptionskip}{-2pt}
\includegraphics[scale=0.35]{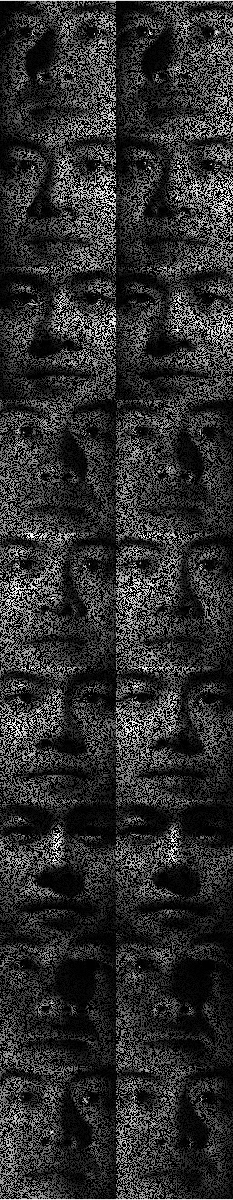}
\subcaption*{RTC-SNN}
\end{subfigure}
\begin{subfigure}[t]{0.09\textwidth}
\centering
\setlength{\abovecaptionskip}{0pt}
\setlength{\belowcaptionskip}{-2pt}
\includegraphics[scale=0.35]{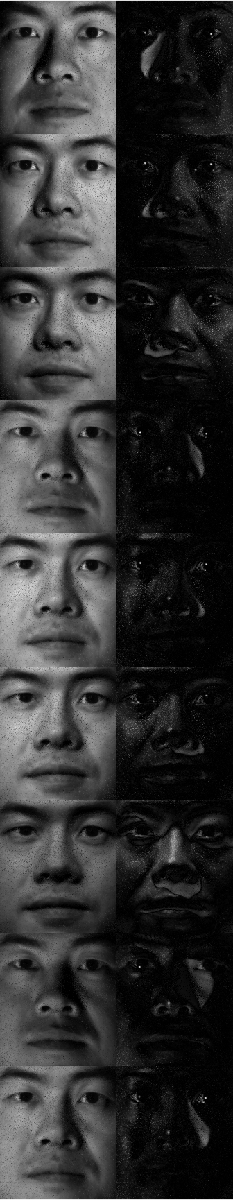}
\subcaption*{RMC}
\end{subfigure}
\begin{subfigure}[t]{0.09\textwidth}
\centering
\setlength{\abovecaptionskip}{0pt}
\setlength{\belowcaptionskip}{-2pt}
\includegraphics[scale=0.35]{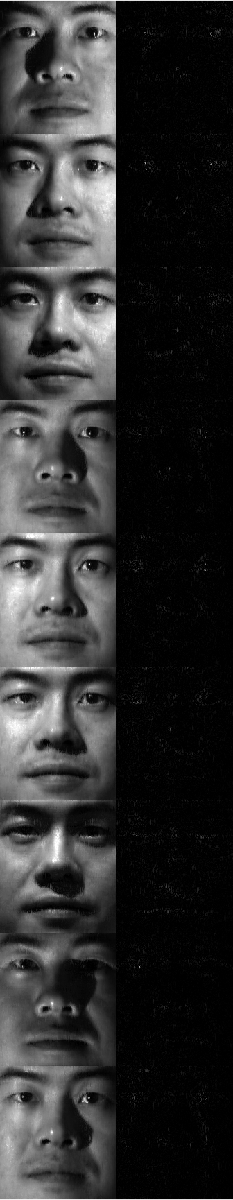}
\subcaption*{SiLRTC-TT}
\end{subfigure}
\begin{subfigure}[t]{0.09\textwidth}
\centering
\setlength{\abovecaptionskip}{0pt}
\setlength{\belowcaptionskip}{-2pt}
\includegraphics[scale=0.35]{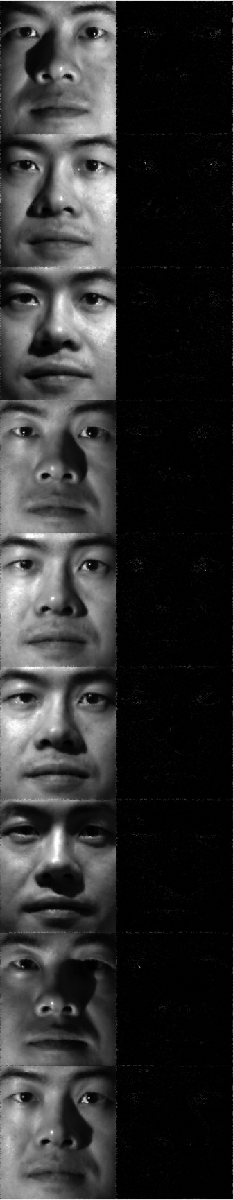}
\subcaption*{STTC}
\end{subfigure}
\begin{subfigure}[t]{0.09\textwidth}
\centering
\setlength{\abovecaptionskip}{0pt}
\setlength{\belowcaptionskip}{-2pt}
\includegraphics[scale=0.35]{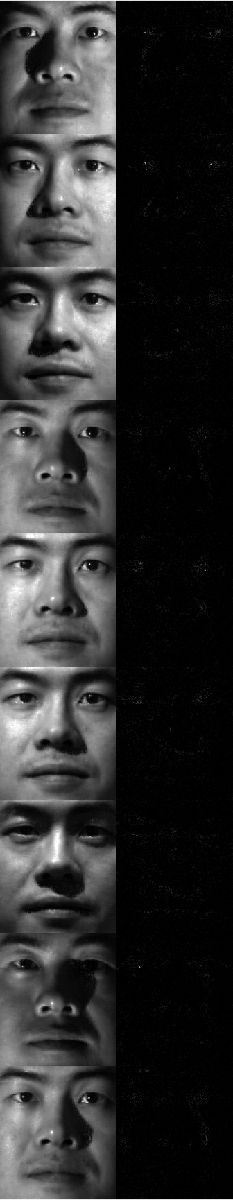}
\subcaption*{TRNNM}
\end{subfigure}
\begin{subfigure}[t]{0.09\textwidth}
\centering
\setlength{\abovecaptionskip}{0pt}
\setlength{\belowcaptionskip}{-2pt}
\includegraphics[scale=0.35]{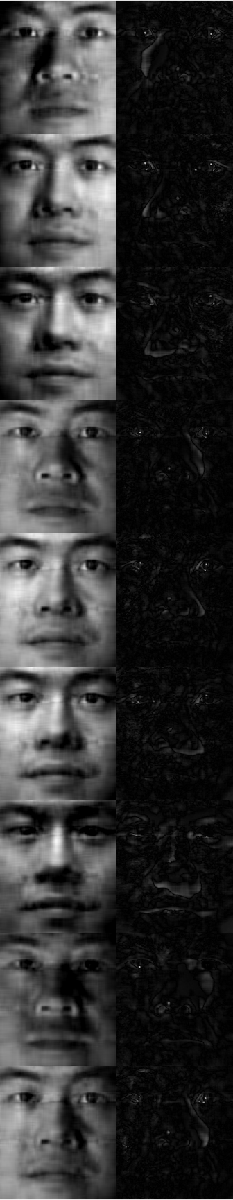}
\subcaption*{FBCP}
\end{subfigure}
\caption{Comparison of eight methods with recovery from the $10$th to $12$th and $27$th-$32$nd frames of subject $1$ and $2$, including RTRC, RTC-TNN, RTC-SNN ,RMC, SiLRTC-TT, STTC, TRNNM and FBCP.  The top nine subfigures are recovery results for subject $1$ and the bottom nine subfigures are recovery results for subject $2$. Each subfigure contains two columns. The subfigures in the first column contain both original images and observations. Other subfigures contain recovery results and absolute differences w.r.t. original images. The averaged time costs are $15.76$s, $95.83$s, $2.42$s, $30.70$s, $7.37$s, $48.12$s, $35.95$s and $89.36$s for the first subject and $13.00$s, $96.89$s, $2.54$s, $27.48$s, $8.48$s, $44.95$s, $34.75$s and $72.02$s for the second subject.}
\label{result-YaleB}
\end{figure*}

The recovery results are provided in Fig. \ref{result-YaleB}. As it shows, the RTRC method eliminates more shadows and costs less time than other methods. In comparison, a few miss entries are not well recovered in the estimates from RMC, and other methods fail to remove the shadow.

\subsection{Color Videos: Application to Background Modeling}
\label{subsection-color video}

\begin{figure}[htbp]
\centering
\begin{subfigure}[t]{0.4\textwidth}
\centering
\setlength{\abovecaptionskip}{0pt}
\setlength{\belowcaptionskip}{-2pt}
\includegraphics[scale=0.27]{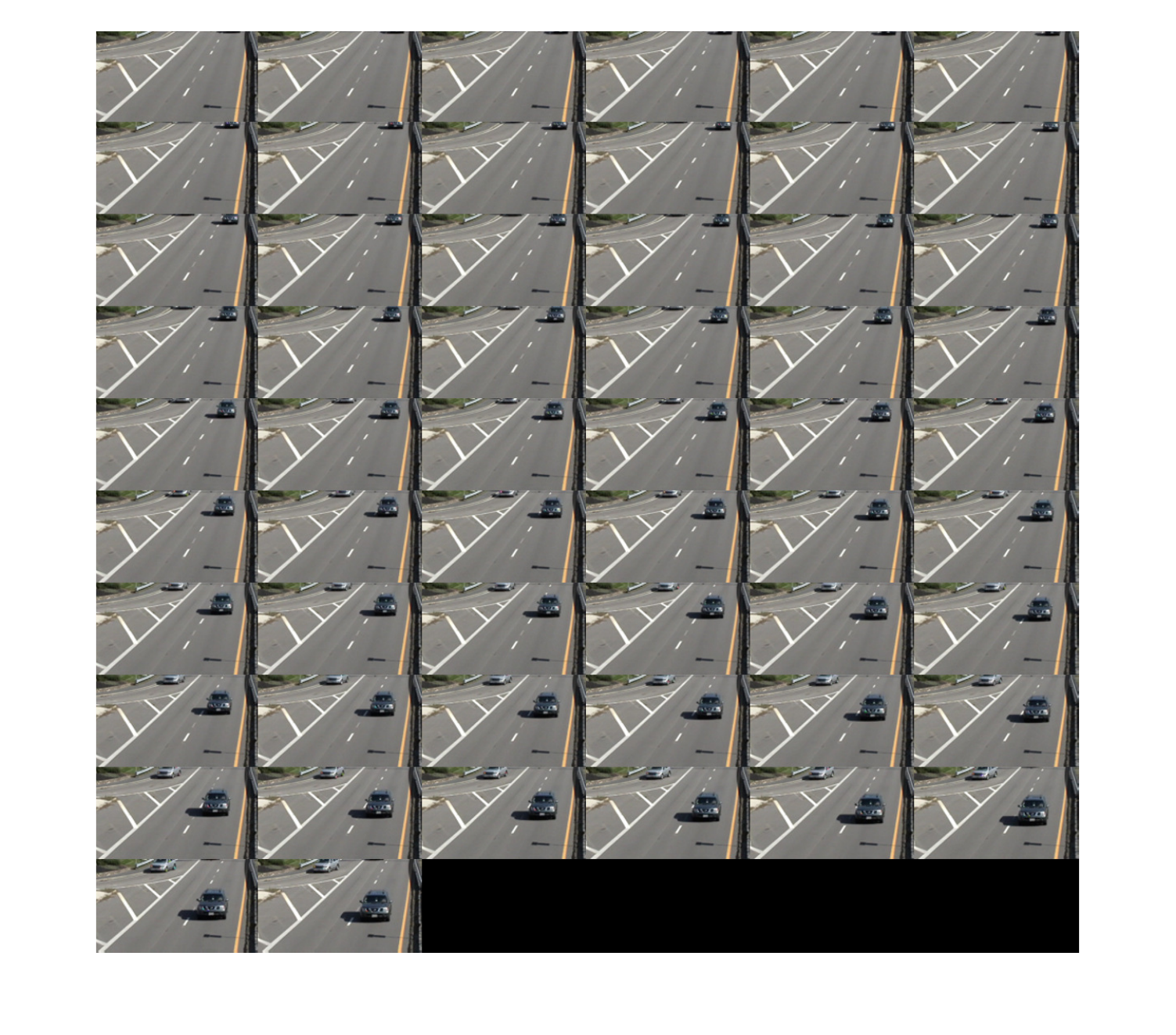}
\subcaption*{\emph{visiontraffic}}
\end{subfigure}

\begin{subfigure}[t]{0.4\textwidth}
\centering
\setlength{\abovecaptionskip}{0pt}
\setlength{\belowcaptionskip}{-2pt}
\includegraphics[scale=0.27]{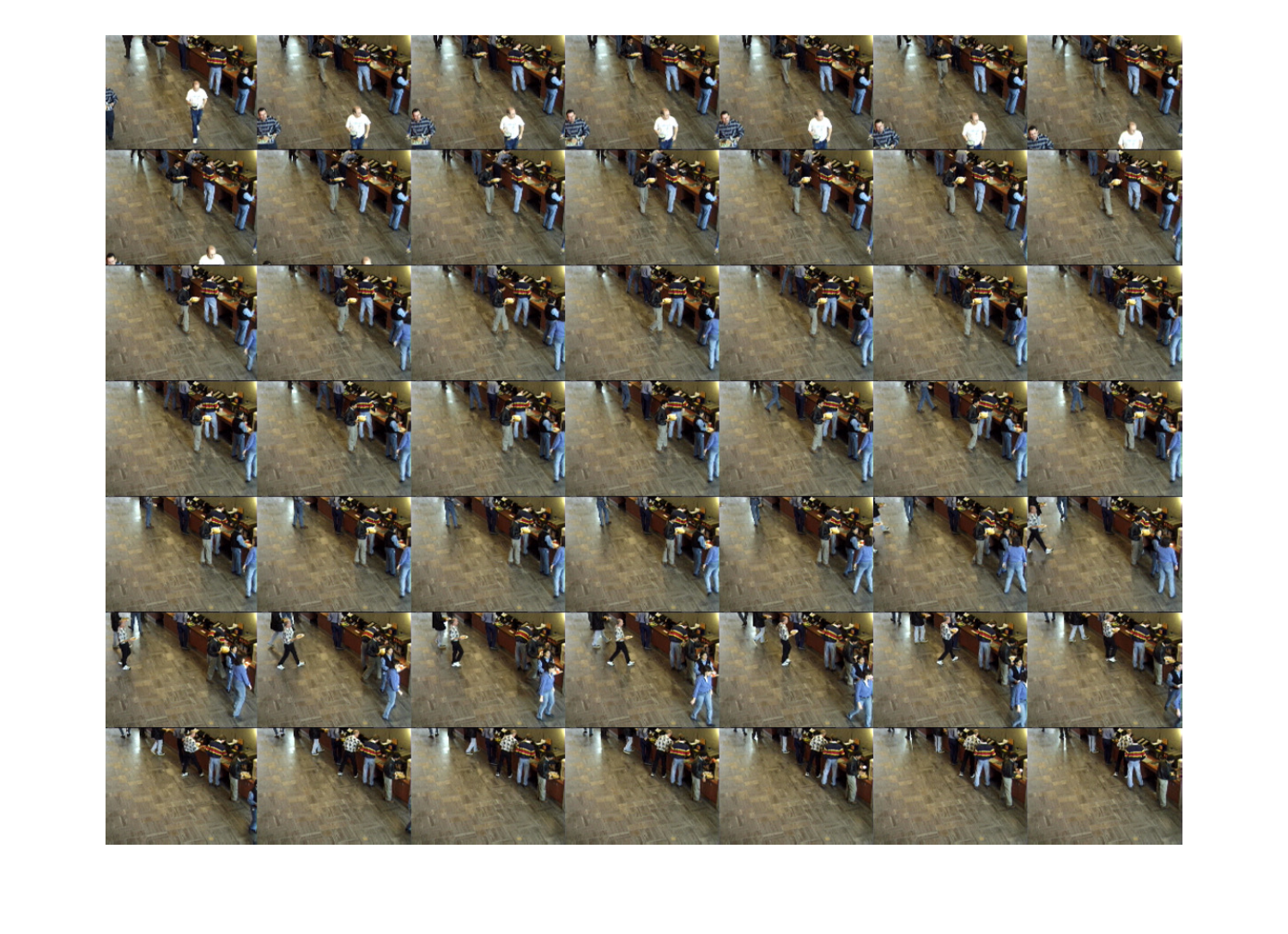}
\subcaption*{\emph{bootstrap}}
\end{subfigure}
\caption{Two color videos.}
\label{video dataset 1}
\end{figure}

In this subsection, two groups of videos are used to test the algorithms. The first color video called \emph{visiontraffic} can be found in MATLAB with size of $288\times352\times3\times531$, and we take its $101$st to $156$th frames. The second video called \emph{bootstrap} comes from the test images for wallflower paper\footnote{https://www.microsoft.com/en-us/download/details.aspx?id=54651}, and we pick up the first $49$ frames which makes a $120\times 160\times 3\times 49$ tensor. The two datasets are shown in Fig. \ref{video dataset 1}. The videos consist of static background and several moving objects which act as foreground components. The background components of these frames are highly correlated, which can be regarded as a low-rank tensor. The foreground occupies a few locations in the whole tensor and plays a role of sparse component. We randomly choose $50\%$ entries as measurements of each dataset and set $10\%$ of observations with their values randomly distributed in $\left[0,255\right]$. Unlike the completion for YaleB fave datasets, this is a more challenging task since there are both uniform and non-uniform sparse noise. To use RTC-TNN, we squeeze the third and the forth dimensions into one dimension. To use RMC, we stack each channel as a column in the matrix. In this group of experiments we set $\alpha=1.1$ and $\beta^0=1\times10^{-4}$ for our methods. Each video recovery is repeated $5$ times to avoid fortuitous result.

\begin{figure*}[htbp]
\centering
\begin{subfigure}[t]{0.09\textwidth}
\centering
\setlength{\abovecaptionskip}{0pt}
\setlength{\belowcaptionskip}{-2pt}
\includegraphics[scale=0.3]{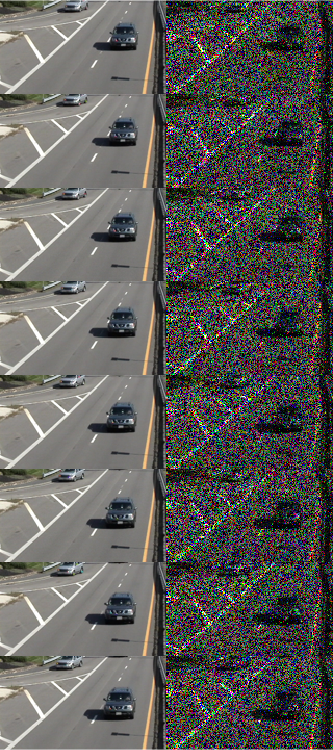}
\subcaption*{\emph{visiontraffic}, observation}
\end{subfigure}
\begin{subfigure}[t]{0.09\textwidth}
\centering
\setlength{\abovecaptionskip}{0pt}
\setlength{\belowcaptionskip}{-2pt}
\includegraphics[scale=0.3]{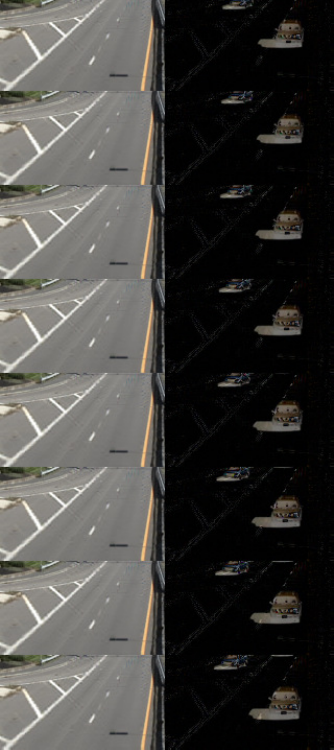}
\subcaption*{RTRC}
\end{subfigure}
\begin{subfigure}[t]{0.09\textwidth}
\centering
\setlength{\abovecaptionskip}{0pt}
\setlength{\belowcaptionskip}{-2pt}
\includegraphics[scale=0.3]{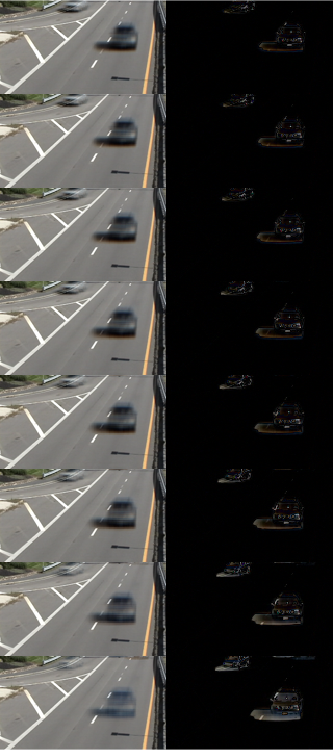}
\subcaption*{RTC-TNN}
\end{subfigure}
\begin{subfigure}[t]{0.09\textwidth}
\centering
\setlength{\abovecaptionskip}{0pt}
\setlength{\belowcaptionskip}{-2pt}
\includegraphics[scale=0.3]{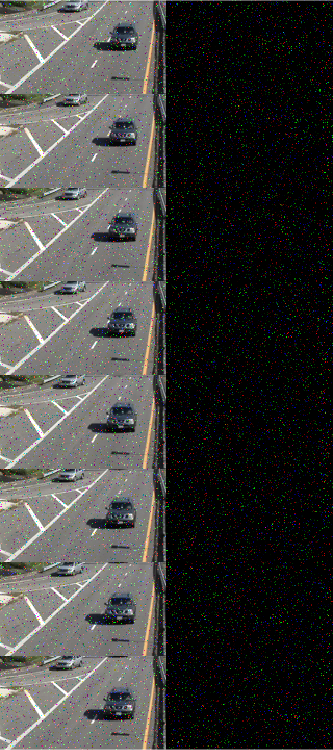}
\subcaption*{RTC-SNN}
\end{subfigure}
\begin{subfigure}[t]{0.09\textwidth}
\centering
\setlength{\abovecaptionskip}{0pt}
\setlength{\belowcaptionskip}{-2pt}
\includegraphics[scale=0.3]{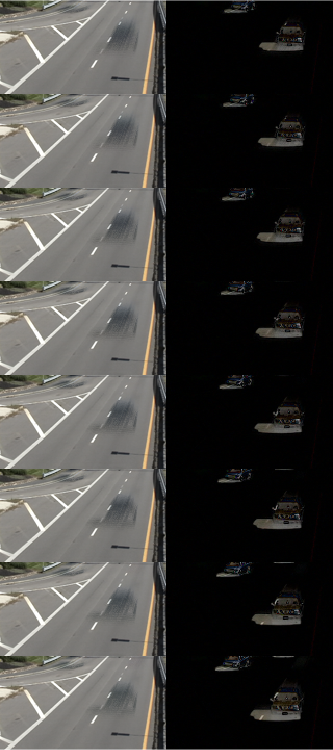}
\subcaption*{RMC}
\end{subfigure}
\begin{subfigure}[t]{0.09\textwidth}
\centering
\setlength{\abovecaptionskip}{0pt}
\setlength{\belowcaptionskip}{-2pt}
\includegraphics[scale=0.3]{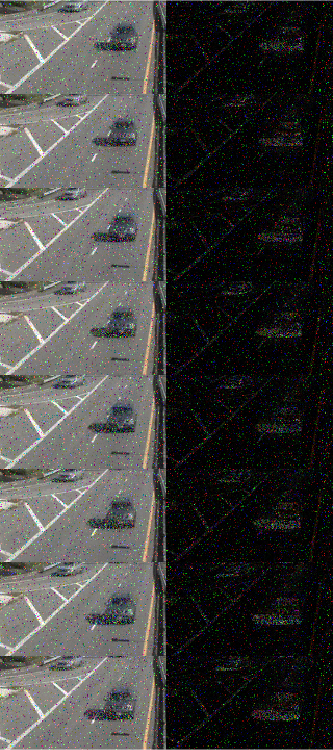}
\subcaption*{SiLRTC-TT}
\end{subfigure}
\begin{subfigure}[t]{0.09\textwidth}
\centering
\setlength{\abovecaptionskip}{0pt}
\setlength{\belowcaptionskip}{-2pt}
\includegraphics[scale=0.3]{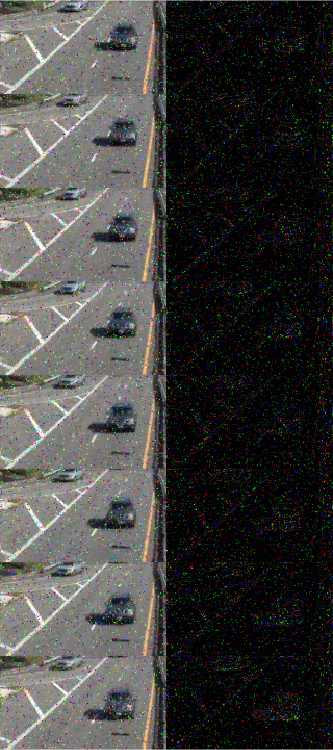}
\subcaption*{STTC}
\end{subfigure}
\begin{subfigure}[t]{0.09\textwidth}
\centering
\setlength{\abovecaptionskip}{0pt}
\setlength{\belowcaptionskip}{-2pt}
\includegraphics[scale=0.3]{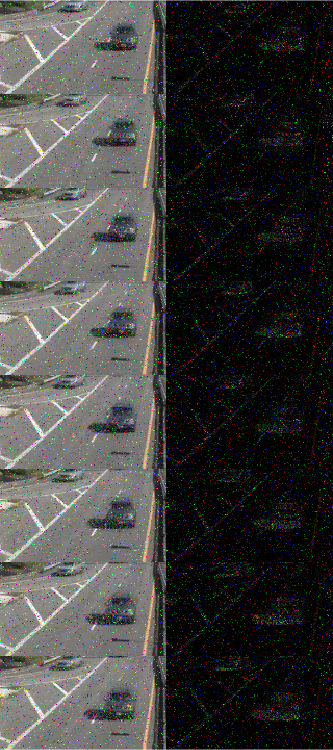}
\subcaption*{TRNNM}
\end{subfigure}
\begin{subfigure}[t]{0.09\textwidth}
\centering
\setlength{\abovecaptionskip}{0pt}
\setlength{\belowcaptionskip}{-2pt}
\includegraphics[scale=0.3]{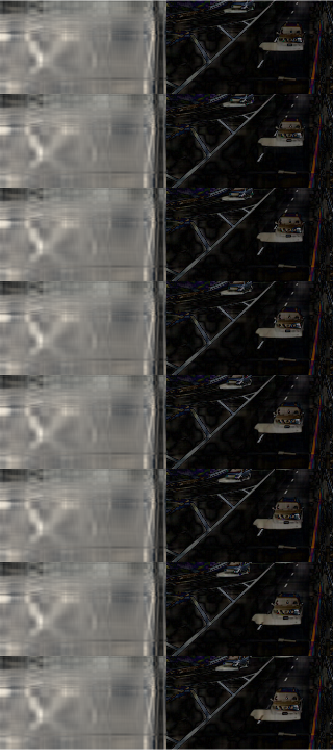}
\subcaption*{FBCP}
\end{subfigure}

\begin{subfigure}[t]{0.09\textwidth}
\centering
\setlength{\abovecaptionskip}{0pt}
\setlength{\belowcaptionskip}{-2pt}
\includegraphics[scale=0.3]{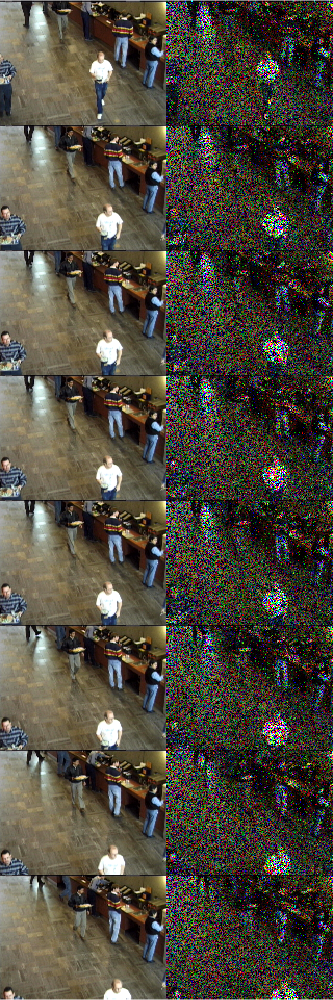}
\subcaption*{\emph{visiontraffic}, observation}
\end{subfigure}
\begin{subfigure}[t]{0.09\textwidth}
\centering
\setlength{\abovecaptionskip}{0pt}
\setlength{\belowcaptionskip}{-2pt}
\includegraphics[scale=0.3]{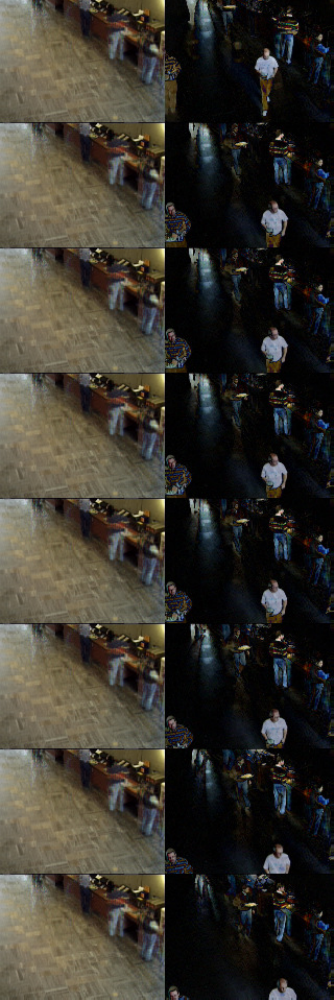}
\subcaption*{RTRC}
\end{subfigure}
\begin{subfigure}[t]{0.09\textwidth}
\centering
\setlength{\abovecaptionskip}{0pt}
\setlength{\belowcaptionskip}{-2pt}
\includegraphics[scale=0.3]{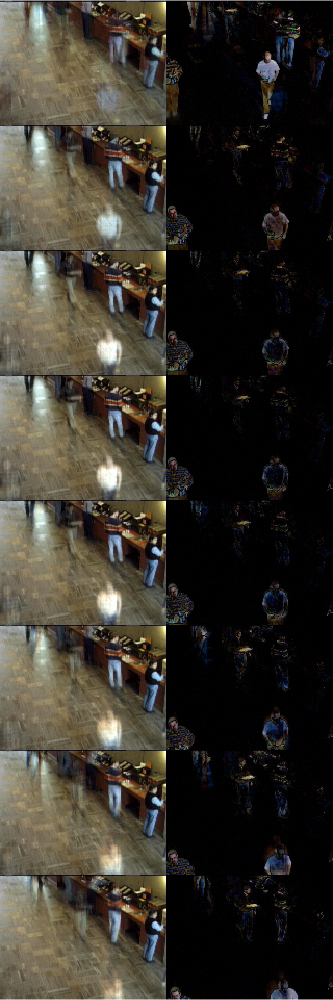}
\subcaption*{RTC-TNN}
\end{subfigure}
\begin{subfigure}[t]{0.09\textwidth}
\centering
\setlength{\abovecaptionskip}{0pt}
\setlength{\belowcaptionskip}{-2pt}
\includegraphics[scale=0.3]{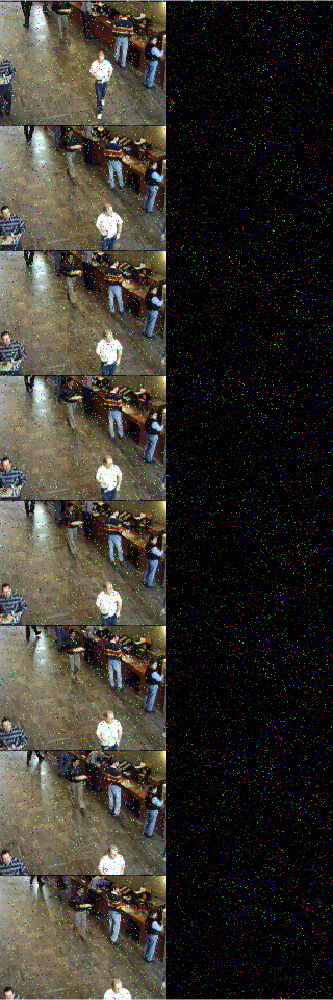}
\subcaption*{RTC-SNN}
\end{subfigure}
\begin{subfigure}[t]{0.09\textwidth}
\centering
\setlength{\abovecaptionskip}{0pt}
\setlength{\belowcaptionskip}{-2pt}
\includegraphics[scale=0.3]{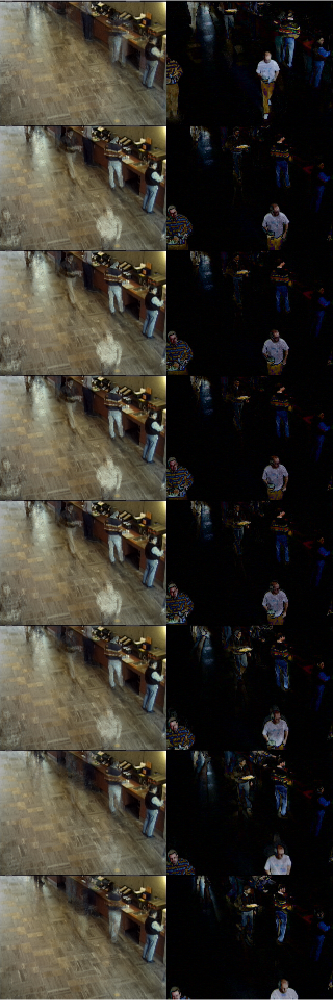}
\subcaption*{RMC}
\end{subfigure}
\begin{subfigure}[t]{0.09\textwidth}
\centering
\setlength{\abovecaptionskip}{0pt}
\setlength{\belowcaptionskip}{-2pt}
\includegraphics[scale=0.3]{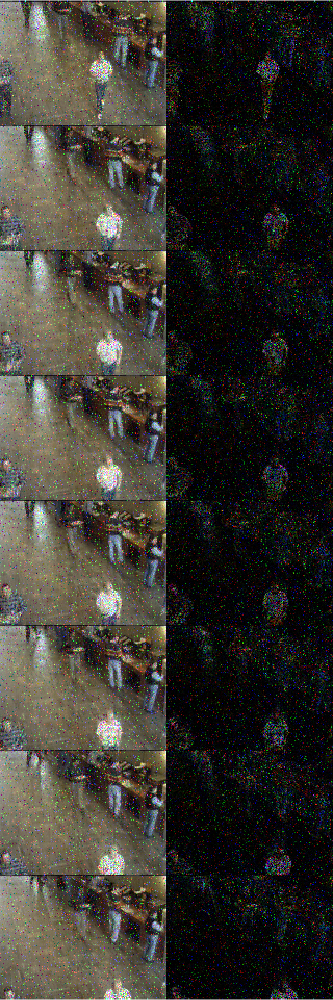}
\subcaption*{SiLRTC-TT}
\end{subfigure}
\begin{subfigure}[t]{0.09\textwidth}
\centering
\setlength{\abovecaptionskip}{0pt}
\setlength{\belowcaptionskip}{-2pt}
\includegraphics[scale=0.3]{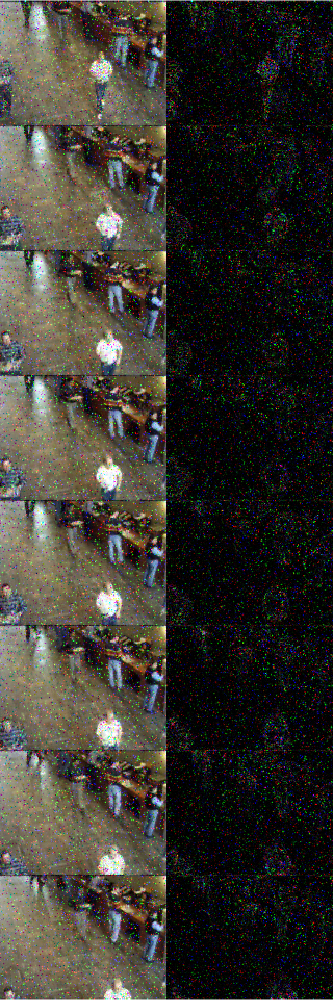}
\subcaption*{STTC}
\end{subfigure}
\begin{subfigure}[t]{0.09\textwidth}
\centering
\setlength{\abovecaptionskip}{0pt}
\setlength{\belowcaptionskip}{-2pt}
\includegraphics[scale=0.3]{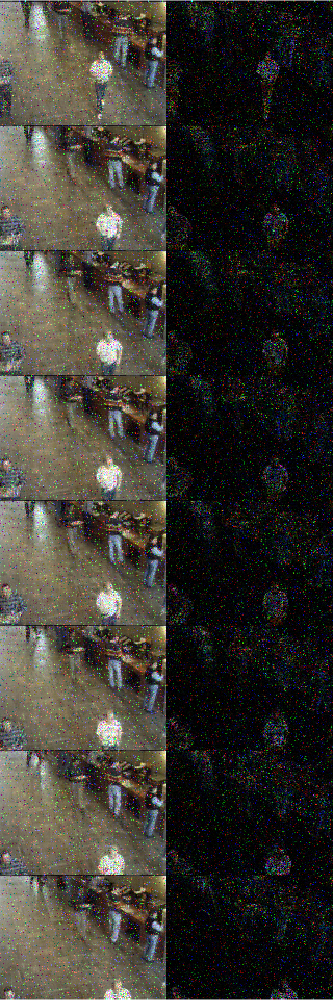}
\subcaption*{TRNNM}
\end{subfigure}
\begin{subfigure}[t]{0.09\textwidth}
\centering
\setlength{\abovecaptionskip}{0pt}
\setlength{\belowcaptionskip}{-2pt}
\includegraphics[scale=0.3]{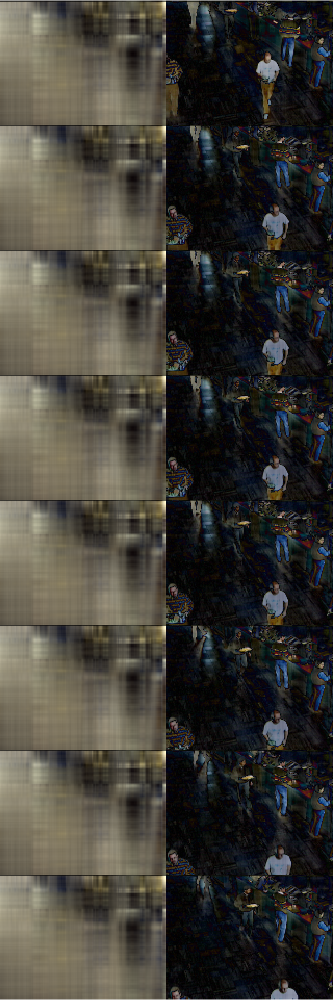}
\subcaption*{FBCP}
\end{subfigure}
\caption{Comparison of eight algorithms with recovery from the last $8$ frames of video \emph{visiontraffic} and first $8$ frames of video \emph{bootstrap}, including RTRC, RTC-TNN, RTC-SNN ,RMC, SiLRTC-TT, STTC, TRNNM and FBCP. The top nine subfigures are recovery results of \emph{visiontraffic} and the bottom nine subfigures are recovery results of \emph{bootstrap}. Each subfigure contains two columns. The  subfigures in the first column contains original images and their observations. The subfigures in other columns are recoveries and absolute differences w.r.t. original ones. The average time costs are $93.90$s, $193.74$s, $7.43$s, $68.15$s $61.34$s, $234.33$s, $147.51$s and $20.18$s for the first video, and $117.63$s, $224.62$s, $7.40$s, $72.11$s, $95.33$s, $226.84$,	$192.77$s and $27.95$s for the second video.}
\label{result-video}
\end{figure*}

Fig. \ref{result-video} gives the results of eight methods that are averaged over $5$ repetitions. The result demonstrates that the RTRC simultaneously separates the background from the foreground and complete the missing entries. It outperforms all other methods in terms of both recovery accuracy and computational cost. The RTC-SNN fails to eliminate the sparse noise. Both RTC-TNN and RMC can recover the missing entries, their background components are not extracted as well as that of RTRC.

\section{Conclusion}
\label{section-conclusion}

In this paper, we propose the robust low rank tensor ring completion. It can be regarded as the extensions of both low rank tensor ring completion and robust tensor principal component analysis based on tensor ring. We rigorously prove  and provide the exact recovery conditions from a partial observation with sparse noise. A number of TR unfoldings are used in the low rank tensor ring term in the optimization model for robust tensor completion, and ADMM is used to solve it. Numerical experiments verifies the recovery condition, and demonstrates that the proposed method  outperforms the state-of-the-art methods in application of visual data recovery, shadow removal and background modeling.

\appendices

\section{Proof of Lemma \ref{lemma1}}
\begin{proof*}
Denote by $\mathbf{A}\in\mathbb{R}^{M\times\prod^{D}_{d=1}N_d}$ the matrix form of linear mapping $\mathscr{A}_{\mathbb{O}}$, where $\mathbb{O}=\left\{j_1,\dotsc,j_M\right\}$ and $\mathbf{A}=\left[\mathbf{e}_{j_1},\dotsc,\mathbf{e}_{j_M}\right]^{\mathrm{T}},\;1\leq j_m \leq M,\;\forall{m}\in\left\{1,\dotsc,M\right\}$. The aforementioned optimization becomes
\begin{equation*}
\min_{\mathcal{A}}\frac{1}{2}{\lVert \mathbf{A}\operatorname{Vec}\left(\mathcal{X}\right)-\mathbf{A}\operatorname{Vec}\left(\mathcal{B}\right) \rVert}^2_2+\tau\lVert \operatorname{Vec}\left(\mathcal{X}\right) \rVert_1,
\end{equation*}
thus its first-order optimality condition is
\begin{equation*}
\mathbf{0}\in\mathbf{A}^{\mathrm{T}}\mathbf{A}\left[\operatorname{Vec}\left(\mathcal{X}\right)-\operatorname{Vec}\left(\mathcal{B}\right)\right]+\tau\partial\lVert \operatorname{Vec}\left(\mathcal{X}\right) \rVert_1,
\end{equation*}
which can be reformulated as
\begin{equation*}
\begin{split}
\mathbf{0}\in& \mathbf{A}^{\mathrm{T}}\mathbf{A}\left[\operatorname{Vec}\left(\mathcal{X}\right)-\operatorname{Vec}\left(\mathcal{B}\right)\right]+\tau\partial\lVert \mathbf{A}^{\mathrm{T}}\mathbf{A}\operatorname{Vec}\left(\mathcal{X}\right) \rVert_1+\\
& \tau\partial\lVert \left(\mathbf{E}-\mathbf{A}^{\mathrm{T}}\mathbf{A}\right)\operatorname{Vec}\left(\mathcal{X}\right) \rVert_1.
\end{split}
\end{equation*}

Define the projection operator $\mathscr{P}_{\mathbb{O}}=\mathscr{A}^*_{\mathbb{O}}\mathscr{A}_{\mathbb{O}}$ and its matrix expression $\mathbf{P}=\mathbf{A}^{\mathrm{T}}\mathbf{A}$, where $\mathscr{A}^*$ represents the adjoint of $\mathscr{A}$. The formula of optimality condition is
\begin{equation*}
\begin{split}
\mathbf{0}\in& \mathscr{P}_{\mathbb{O}}\left(\operatorname{Vec}\left(\mathcal{X}\right)-\operatorname{Vec}\left(\mathcal{B}\right)\right)+\tau\partial\lVert \mathscr{P}_{\mathbb{O}}\left(\operatorname{Vec}\left(\mathcal{X}\right)\right) \rVert_1+\\
& \tau\partial\lVert \mathscr{P}_{\mathbb{O}^\perp}\left(\operatorname{Vec}\left(\mathcal{X}\right)\right) \rVert_1.
\end{split}
\end{equation*}
In order to minimize the $\ell_1$ norm, the value of component under projection $\mathbb{O}^\perp$ should be $0$. Note that the projection satisfies $\mathbf{A}^{\mathrm{T}}\mathbf{A}=\sum^{m}_{i=1}\mathbf{e}_{j_i}\mathbf{e}^{\mathrm{T}}_{j_i}=\operatorname{diag}\left(\dotsc,j_1,0,\dotsc,j_m,0,\dotsc\right)=\operatorname{diag}\left(\operatorname{Vec}\left(\mathcal{P}\right)\right)$. Rewrite the condition as
\begin{equation*}
\begin{split}
\mathbf{0}\in& \mathcal{P}\circledast\mathcal{X}-\mathcal{P}\circledast\mathcal{B}+\tau\partial\lVert \mathcal{P}\circledast\mathcal{X} \rVert_1,
\end{split}
\end{equation*}
which is also the optimality condition of
\begin{equation*}
\min_{\mathcal{A}}\frac{1}{2}{\lVert \mathcal{P}\circledast\mathcal{X}-\mathcal{P}\circledast\mathcal{B} \rVert}^2_{\mathrm{F}}+\tau\lVert \mathcal{P}\circledast\mathcal{X} \rVert_1.
\end{equation*}
Since $\operatorname{sgn}\left(0\right)=0$, the optimal solution is given by $\mathcal{X}^*=\operatorname{S}_{\tau}\left(\mathcal{P}\circledast\mathcal{B}\right)$.

$\hfill \blacksquare$
\end{proof*}



\ifCLASSOPTIONcaptionsoff
  \newpage
\fi



%



\bibliographystyle{ieeetr}
\bibliography{references_RTRC}

\end{document}